\documentclass[lettersize,journal]{IEEEtran}
\IEEEoverridecommandlockouts
\usepackage{pifont} % 182/183/184
\usepackage{subcaption}
\usepackage{verbatim}
\usepackage{multirow}
\usepackage{xspace}
\usepackage{amsfonts,amssymb} 
\usepackage{microtype}
\usepackage{amsmath}
\usepackage{amsthm}
\usepackage{booktabs}
\usepackage{latexsym}
\usepackage[dvipsnames, table]{xcolor} 
\usepackage{soul}
\usepackage{makecell}  
\usepackage{tabularx}  
\usepackage{adjustbox}
\usepackage{listings}
\usepackage{bbding}
\usepackage{wasysym}
\usepackage{utfsym}
\usepackage{fontawesome}
\usepackage{mdframed}
\usepackage{cite}
\usepackage{textcomp}
\usepackage{threeparttable}
\def\BibTeX{{\rm B\kern-.05em{\sc i\kern-.025em b}\kern-.08em
    T\kern-.1667em\lower.7ex\hbox{E}\kern-.125emX}}
\definecolor{color_gray}{RGB}{229,229,229}
\definecolor{color_pink}{RGB}{252,182,165}
\definecolor{color_orange}{RGB}{255,217,178}
\definecolor{color_yellow}{RGB}{255,255,204}
\definecolor{color_blue}{RGB}{189,215,238}
\cellcolor{color_blue} \cellcolor{color_pink} \cellcolor{color_yellow} \cellcolor{blue!8}

\begin{document}

\title{FedCoE: Bridging Generalization and Personalization via Federated Coordinated Dual-level MoEs
}

\author{Penglin~Dai,~\IEEEmembership{Member,~IEEE}, Fulian~Li, Xincao~Xu,~\IEEEmembership{Member,~IEEE,}
Junhua~Wang,~\IEEEmembership{Member,~IEEE,}\\
Lixin Duan,~\IEEEmembership{Member,~IEEE,}
Xiao~Wu,~\IEEEmembership{Member,~IEEE}
        % <-this % stops a space
\thanks{This work was supported in part by the National
Natural Science Foundation of China under Grant 62172342; in part by the Guangdong Basic and Applied Basic Research Foundation under Grant 2025A1515012825; in part by Yibin Science and Technology Program under Grant 2025JC014, and in part by Science, Technology and Innovation Project of Shenzhen Longhua District under Grant 20260309G23410662. }
\thanks{P. Dai, F. Li, X. Wu are with the School of Computing and Artificial Intelligence, Southwest Jiaotong University, Chengdu 611756, China. (email: penglindai@swjtu.edu.cn; fulianli@my.swjtu.edu.cn; wuxiaohk@swjtu.edu.cn).}%
\thanks{X. Xu, L, Duan are with the Shenzhen Institute for Advanced Study, University of Electronic Science and Technology of China, Shenzhen 518110, China. (email: xc.xu@uestc.edu.cn; lxduan@uestc.edu.cn.)}
\thanks{J. Wang is with the School of Computer Science and Engineering, Northeastern University, Shenyang 110819, China. (email: wangjunhua@cse.neu.edu.cn)}}
\maketitle

\begin{abstract}
Federated Learning (FL) has emerged as a promising paradigm for privacy-preserving distributed learning. However, existing FL methods face a fundamental challenge. Traditional averaging-based approaches suffer from parameter divergence under non-IID conditions, while personalized FL methods overfit to local data and fail to generalize to new clients (cold-start problem). Mixture-of-Experts naturally addresses this by routing heterogeneous data to specialized experts rather than forcing uniform aggregation. In this paper, we propose FedCoE, a \textbf{Fed}erated \textbf{Co}ordinated dual-level mixture-of-\textbf{E}xperts framework that effectively balances global generalization with local personalization. FedCoE maintains multiple independent global expert models on the server and employs a shared gating network to dynamically model client-expert correlations during aggregation, effectively mitigating expert drift and gating inconsistency. To address the cold-start challenge, we introduce an adaptive mechanism that enables new clients to immediately leverage the global expert pool without extensive local training. Extensive experiments demonstrate that FedCoE achieves 78.00\% global accuracy and 89.32\% personalized accuracy on average, outperforming the baseline by 8.82\% and 29.19\%, respectively. In cold-start scenarios, FedCoE delivers 77.27\% accuracy without any local fine-tuning, outperforming baselines by over 12.54\%.
\end{abstract}

\begin{IEEEkeywords}
Federated Learning, Mixture of Experts, Personalized Federated Learning, Non-IID Data, Gold-Start Problem 
\end{IEEEkeywords}

\section{Introduction}
The unprecedented success of deep learning relies on the availability of large-scale data \cite{lecun2015deep}.
Yet in practice, data is fragmented across isolated institutions and devices, each bound by privacy regulations such as GDPR \cite{regulation2018general}, CCPA \cite{pardau2018california} that prohibit sharing. 
This has led to the emergence of ``data silos,'' where valuable data remains locked in local storage, inaccessible for global model training \cite{kairouz2021advances}.
Federated Learning (FL) \cite{mcmahan2017communication} was introduced to resolve this tension, enabling collaborative model training without exposing raw data.
Although FL methods based on averaging aggregation, such as FedAvg \cite{mcmahan2017communication} and FedProx \cite{li2020federated}, have achieved success in many scenarios, they face a fundamental dilemma under non-IID conditions. 
On one hand, the heterogeneity of distributed data is precisely what makes it valuable, as diverse sources provide broader coverage and richer patterns. 
On the other hand, this same heterogeneity causes parameter divergence and unstable convergence, degrading global model performance \cite{li2022federated, xu2024overcoming}. 
In other words, the very property that makes distributed data worth aggregating is also what makes aggregation fail.

% The unprecedented success of deep learning in domains such as computer vision and natural language processing is predominantly driven by the availability of large-scale, centralised datasets \cite{lecun2015deep}. However, the increasing enforcement of rigorous privacy regulations (e.g., GDPR \cite{regulation2018general}, CCPA \cite{pardau2018california}) and the inherent fragmentation of data across isolated devices and institutions have rendered centralized data collection increasingly impractical. This has led to the emergence of ``data silos,'' where valuable data remains locked in local storage, inaccessible for global model training \cite{kairouz2021advances}.
% Federated Learning (FL) \cite{mcmahan2017communication} has been proposed to utilize these distributed data resources while maintaining data privacy. It allows multiple clients to upload only their model updates to a central server for aggregation, avoiding the transmission of any raw data. This mechanism effectively balances privacy protection with collaborative model training across different data sources.
% Although federated learning methods based on averaging aggregation, such as FedAvg\cite{mcmahan2017communication} and FedProx \cite{li2020federated}, have been successful in many scenarios, they suffer under non-IID conditions, where client data distributions differ significantly. Such heterogeneity can cause parameter divergence, unstable convergence, and degraded global performance \cite{li2022federated,xu2024overcoming}.

To ameliorate the restrictiveness of traditional FL, Personalized Federated Learning (pFL) has been introduced to reconcile global collaboration with local adaptation \cite{arivazhagan2019federated,wang2023towards}. Conventional pFL strategies typically decouple the model into shared and personalized modules, allowing the former to capture universal features while the latter adapts to client-specific distributions. However, this localization comes at a critical cost: by prioritizing client-specific optimization, these methods often induce \textit{severe overfitting} to local distributions, thereby degrading the generalizability of the global model. A common failure mode is that the aggregated model loses its ability to represent the broader data manifold, becoming a fragmented collection of specialized parameters rather than a cohesive global learner. This deficiency is particularly pronounced in the \textit{cold-start} scenario, where new clients entering the federation receive a biased initialization that is ill-suited to their unseen distributions, resulting in prohibitive fine-tuning costs and poor initial performance. 
This reveals a fundamental \textit{trilemma} in federated learning, where global generalization, local personalization, and cold-start robustness appear mutually exclusive under existing frameworks. 
Addressing this trilemma requires a new architectural paradigm.
% Thus, achieving a harmonious balance between high-precision personalization and robust global generalization remains a persistent challenge.

% In pursuit of this balance, researchers have increasingly investigated architectures capable of dynamic adaptation across heterogeneous distributions, with the Mixture of Experts (MoE) emerging as a particularly promising candidate. Characterized by its sparse activation and parameter efficiency, MoE has seen widespread adoption in large-scale pre-trained models \cite{shazeer2017outrageously,lepikhin2020gshard}. A canonical MoE system comprises a gating network and a set of internal experts. The gating network dynamically computes weighting coefficients based on input features, selectively activating only the top-$k$ most relevant experts for parallel computation. This conditional computation mechanism not only scales model capacity without proportional increases in computational cost \cite{fedus2022switch,riquelme2021scaling} but also intrinsically fosters a synergy between generalization and specialization. By routing inputs to the most suitable experts, the model can learn diverse data patterns within a unified parameter space \cite{lepikhin2020gshard,zoph2022designing}, where individual experts specialize in distinct feature domains while the gating network maintains global coherence. Such structured representations offer a compelling avenue for addressing the statistical heterogeneity inherent in federated learning \cite{rajbhandari2022deepspeed}.

The Mixture of Experts (MoE) architecture offers a promising solution to this dilemma \cite{shazeer2017outrageously, lepikhin2020gshard, rajbhandari2022deepspeed}. 
By dynamically routing inputs to specialized experts rather than forcing uniform aggregation, MoE naturally accommodates heterogeneous data distributions while maintaining a unified model structure \cite{lepikhin2020gshard,zoph2022designing}. 
This conditional computation mechanism enables individual experts to specialize in distinct domains while the gating network coordinates global knowledge sharing \cite{fedus2022switch, riquelme2021scaling}.
However, directly deploying MoE within FL introduces three critical challenges.
First, \textit{Expert Drift} emerges as a primary consequence of statistical heterogeneity. Since clients optimize subsets of experts solely on disjoint local distributions, the aggregated global experts often suffer from divergent semantic representations and functional redundancy, degrading the coherence of the global model \cite{zou2024fed,mei2024fedmoe}.
Second, \textit{Gating Inconsistency} severely undermines convergence. Trained in isolation on local data manifolds, client-side gating networks inevitably develop discordant routing policies. This lack of consensus prevents the establishment of a unified expert selection strategy, thereby impairing the global model's ability to generalise across diverse clients \cite{wu2024fedmoe,liang2025mixture}.
Third, \textit{Resource Constraints} present a prohibitive bottleneck. The naive exchange of full expert ensembles imposes substantial communication overhead and computational burdens, negating the efficiency gains of sparse activation. 
Addressing these challenges requires rethinking how experts are coordinated across the federation, effectively reconciling personalized adaptation with robust global generalization.
% Collectively, these challenges prevent existing MoE-FL frameworks from effectively reconciling personalized adaptation with robust global generalization.

To surmount these impediments, we introduce \textbf{FedCoE}, a \textbf{Fed}erated \textbf{Co}ordinated dual-level mixture-of-\textbf{E}xperts framework designed to enforce semantic consistency across the federation. FedCoE fundamentally re-engineers the MoE-FL paradigm through three targeted innovations corresponding to the aforementioned challenges. First, to counter \textit{Expert Drift}, we develop a correlation-based aggregation mechanism that selectively updates global experts using only semantically aligned client gradients, thereby preserving functional specialization. Second, to resolve \textit{Gating Inconsistency}, we deploy a shared server-side gating network that acts as a global anchor, synchronizing routing policies across heterogeneous clients to ensure coherent expert selection. Third, to alleviate \textit{Resource Constraints} and \textit{Cold-Start} latency, we introduce an adaptive expert assembly strategy: new clients utilize the lightweight gating network to instantly pinpoint and retrieve a personalized expert subset, achieving robust initialization without the communication burden of full model transfer or the computational cost of iterative fine-tuning.

In summary, the main contributions of this work are listed as follows:
\begin{itemize}
\item[$\bullet$] We propose FedCoE, a new federated coordinated dual-level mixture-of-experts framework that redefines the MoE-FL integration through a dual-level architecture where both server and clients share a homogeneous MoE backbone. Unlike prior works that treat MoE primarily as a compression or efficiency tool, our framework leverages it as a semantic coordination mechanism for bridging global generalization and local personalization.
% We propose a unified dual-expert framework that effectively synergizes global generalization with local personalization. By enforcing semantic consistency through a homogeneous MoE architecture across the server and clients, we resolve the structural divergence limitations inherent in conventional federated learning methods.
\item[$\bullet$] We present a consistency-driven expert aggregation strategy driven by a shared semantic gating network that computes client-expert correlation matrices, enabling selective knowledge absorption from semantically aligned distributions. This design directly addresses the critical issues of expert drift and gating inconsistency that undermine existing MoE-FL methods.
% We design a correlation-aware aggregation mechanism driven by a shared gating network. This innovation directly mitigates the critical issues of expert drift and gating inconsistency by ensuring that global experts selectively absorb knowledge from semantically aligned data distributions.
\item[$\bullet$] We propose an adaptive cold-start mechanism that enables new clients to immediately obtain personalized models by profiling their local distribution through the shared gating network, completely bypassing expensive fine-tuning. To the best of our knowledge, this is the first MoE-based FL method to achieve genuine zero-shot initialization for unseen clients.
\item[$\bullet$] We conduct extensive experiments across IID and various non-IID settings, demonstrating that FedCoE achieves state-of-the-art performance in both global generalization and local personalization. Notably, FedCoE achieves 78\% global accuracy and 89.32\% personalized accuracy across various non-IID settings, outperforming the baseline by 8.82\% and 29.19\%, respectively. More importantly, new clients can attain 77.27\% accuracy through zero-shot expert assembly alone, surpassing the baseline by over 12.54\% and demonstrating the practical effectiveness of our cold-start solution.
% We conduct extensive experiments across various non-IID settings, demonstrating that FedCoE achieves superior performance with \textbf{85.55\%} global accuracy and \textbf{90.26\%} personalized accuracy. Furthermore, our results validate that the proposed adaptive expert assembly strategy effectively solves the cold-start problem, enabling robust zero-shot initialization for new clients.
\end{itemize}

The remainder of this paper is organized as follows. Section \ref{sec:related_work} reviews the related literature. Section \ref{sec:motivation} outlines the motivation and preliminary experiment. Section \ref{sec:alg} details the proposed FedCoE framework. Section \ref{sec:performance_evaluation} presents the experimental results and analysis. Finally, Section \ref{sec:conclusion} concludes the paper.

\section{Related Work}\label{sec:related_work}

\begin{table*}[t]
\centering
\caption{Comparison of FedCoE with existing representative federated learning methods. ($\checkmark$: Supported/Good, $\triangle$: Partially Supported/Limited, $\times$: Not Supported/Poor)}
\label{tab:related_work_comparison}
% \resizebox{0.95\textwidth}{!}{%
\footnotesize
\begin{tabular}{llcccc}
\toprule
\textbf{Category} & \textbf{Method} & \textbf{Core Strategy} & \textbf{Personalization} & \textbf{Semantic/Gating Consistency} & \textbf{Cold-Start Capability} \\ 
\midrule
\multirow{3}{*}{Traditional FL} 
& FedAvg \cite{mcmahan2017communication} & Parameter Averaging & $\times$ & $\times$ & $\times$ \\
& FedProx \cite{li2020federated} & Regularization & $\times$ & $\times$ & $\times$ \\
& SCAFFOLD \cite{karimireddy2020scaffold} & Control Variates & $\times$ & $\times$ & $\times$ \\ 
\midrule
\multirow{3}{*}{Personalized FL} 
& FedPer \cite{arivazhagan2019federated} & Layer Splitting & $\checkmark$ & $\times$ & $\times$ \\
& FedFomo \cite{zhangpersonalized} & Similarity Weighting & $\checkmark$ & $\triangle$ (Geometric) & $\times$ \\
& FedKD \cite{wu2022communication} & Knowledge Distillation & $\checkmark$ & $\triangle$ (Soft Labels) & $\times$ \\ 
\midrule
\multirow{4}{*}{MoE-based FL} 
& FedMoE \cite{mei2024fedmoe} & Task-Specific Update & $\times$ & $\times$ & $\times$ \\
& Fed-MoE \cite{zou2024fed} & Pruning \& Merging & $\triangle$ & $\triangle$ & $\times$ \\
& FedJETs \cite{dunfedjets} & Dynamic Gating & $\checkmark$ & $\times$ & \textbf{$\checkmark$} \\
& \textbf{FedCoE (Ours)} & \textbf{Correlation-aware Dual-MoE} & \textbf{$\checkmark$} & \textbf{$\checkmark$} & \textbf{$\checkmark$} \\ 
\bottomrule
\end{tabular}%
% }
\end{table*}

\subsection{Federated Learning and Data Heterogeneity}
The foundational algorithm, FedAvg \cite{mcmahan2017communication}, aggregates a global model by averaging local updates. However, it suffers from severe client drift under non-IID conditions. To mitigate this, FedProx \cite{li2020federated} introduces a proximal term to the local objective, penalizing deviations from the global model to enforce stability. SCAFFOLD \cite{karimireddy2020scaffold} further addresses drift by utilizing control variates to correct the direction of local gradients. Despite these improvements, recent studies \cite{li2022federated,xu2024overcoming} show that constraining local updates often compromises the model's ability to capture diverse data distributions, leading to suboptimal generalization.
More advanced approaches employ weighted aggregation. For instance, FedFomo \cite{zhangpersonalized} calculates aggregation weights based on the geometric similarity between local and global models. Similarly, FedKD \cite{wu2022communication} uses knowledge distillation to align client outputs. However, these methods rely heavily on superficial statistical or geometric metrics, lacking the deep semantic modeling required to handle complex, highly heterogeneous data effectively.

\subsection{Personalized Federated Learning}
Personalized FL (pFL) tailors models to individual clients. A representative strategy is architectural splitting, as seen in FedPer \cite{arivazhagan2019federated} and LG-FedAvg \cite{liang2020thinklocallyactglobally}, and recent architecture search approaches \cite{yao2024perfedrlnas}, where the model is decoupled into shared base layers (for global features) and private personalization layers (for local adaptation). While effective for local tasks, this decoupling often leads to a fragmented global model that fails to generalize across clients. Another direction involves meta-learning, such as Per-FedAvg \cite{fallah2020personalized}and PeFLL \cite{scott2024pefll}, which seeks an initialization point that can be quickly adapted to local tasks via fine-tuning. However, these methods typically require computationally expensive local gradient steps for new clients, making them unsuitable for rapid cold-start scenarios where data is scarce or instant inference is required.

\subsection{Mixture-of-Experts in Federated Learning}
\subsubsection{Global Federated MoE}
Integrating Mixture-of-Experts (MoE) into FL leverages sparse activation for efficiency. FedMoE \cite{mei2024fedmoe} allows clients to update only task-relevant experts, reducing communication costs. Fed-MoE \cite{zou2024fed} constructs a global expert pool via similarity-based pruning and merging. While these methods improve training efficiency, they treat MoE primarily as a compression or resource-optimization tool rather than a mechanism for disentangling complex data distributions.

\subsubsection{MoE-Driven Personalization}
Recent works exploit MoE for personalization. Fed-POE \cite{m2024personalized} and FedJETs \cite{dunfedjets} employs a gating network to select experts for each client, but it aggregates experts via simple averaging, which dilutes specialized knowledge. PM-MoE \cite{feng2025pm} attempts to fuse personalized modules using gating weights but suffers from coarse granularity, operating at the client level rather than the sample level. FedMoEKD \cite{liang2025mixture} introduces distillation but relies on unstable similarity metrics for expert weighting. Hetero-FedMoE \cite{jiang2025heterogeneous} introduces a mixture of experts framework specifically designed to mitigate client drift caused by non-IID data distributions. Crucially, these methods lack a mechanism to synchronize the routing logic across clients, leading to \textit{gating inconsistency} where different clients develop conflicting expert selection policies.

\textit{Difference Summary:} The proposed \textbf{FedCoE} distinguishes itself from existing works through three key innovations. First, unlike methods that rely on geometric similarity (e.g., FedFomo) or simple averaging (e.g., FedJETs), FedCoE introduces a \textbf{correlation-aware aggregation} mechanism that updates experts based on deep semantic alignment, preserving their functional specialization. Second, addressing the \textit{gating inconsistency} prevalent in prior arts (e.g., PM-MoE), FedCoE enforces a unified routing policy via a \textbf{shared server-side gating network}, ensuring coherent expert selection across the federation. Third, in contrast to meta-learning approaches requiring local fine-tuning (e.g., Per-FedAvg), FedCoE offers a genuine \textbf{zero-shot cold-start solution} by utilizing the shared gating network to instantly assemble a personalized expert model for new clients, completely bypassing the need for local gradient updates.

\section{Motivation and Preliminary Experiment}\label{sec:motivation}
To investigate whether MoE can bridge the generalization-personalization gap in federated learning, we conduct preliminary experiments comparing ResNet-MoE \cite{zhang2023robust} with standard ResNet under both IID and non-IID settings as shown in Fig.\ref{fig:ob1}. Three key observations emerge:

\begin{figure}[t]
    \centering
    \includegraphics[width=1\linewidth]{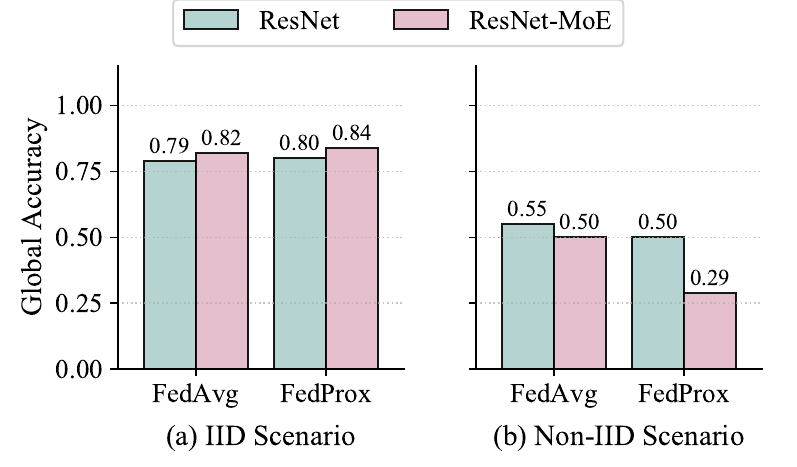}
    \caption{Performance comparison of ResNet and ResNet-MoE}
    \label{fig:ob1}
\end{figure}

\begin{figure}[t]
    \centering
    \includegraphics[width=1\linewidth]{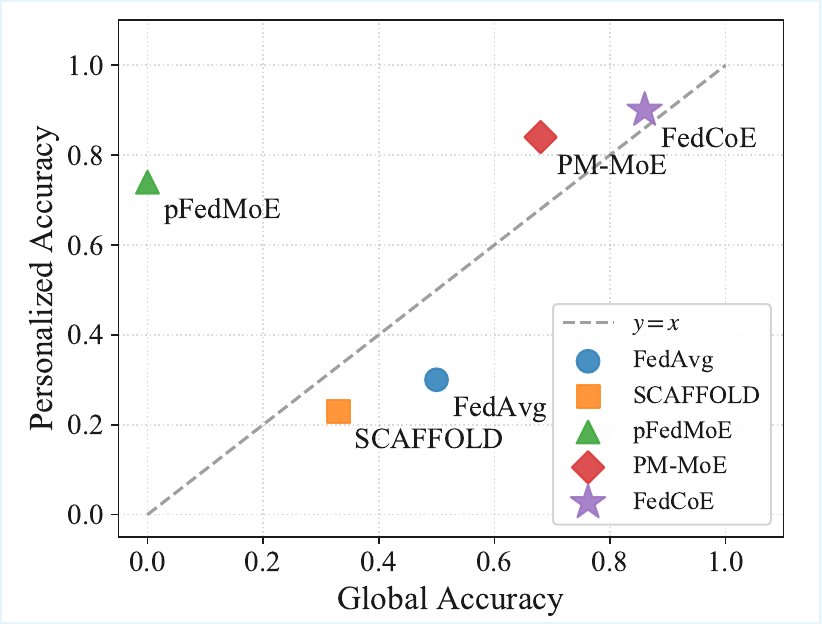}
    \caption{Global vs. Personalized Accuracy Comparison. \textit{*Note: Global Accuracy for pFedMoE is plotted as 0 as they lack a unified global model.}}
    \label{fig:ob3}
\end{figure}

% Mixture-of-Experts (MoE) architectures have recently demonstrated remarkable scalability and performance in large-scale deep learning models. By employing a sparse activation mechanism where a routing network dynamically selects a subset of experts for each input, MoE achieves a superior balance between model capacity and computational efficiency. In this work, we adopt ResNet-MoE \cite{zhang2023robust} as our backbone architecture (Fig.\ref{fig:moe}). This design integrates multiple lightweight convolutional experts within each residual block, managed by a router that activates only the top-$k$ most relevant experts. This capability to decouple feature learning into specialized subspaces offers a promising avenue for Federated Learning, potentially enabling simultaneous global generalization and local personalization. To validate this potential and identify critical challenges, we conducted preliminary experiments comparing ResNet-MoE with standard ResNet under federated settings.

\textbf{Observation 1: MoE Excels in IID Settings.} Under IID conditions, ResNet-MoE achieves 81.51\% accuracy compared to 78.93\% for standard ResNet, a 2.58\% improvement as shown in Fig.\ref{fig:ob1}(a). The sparse expert mechanism enables diverse pattern learning within a unified parameter space, effectively enhancing representation power and convergence efficiency without incurring excessive computational costs, and validating MoE's potential for federated scenarios.
% \textbf{Observation 1: MoE Excels in IID Settings.} Under independent and identically distributed (IID) conditions, the MoE architecture significantly outperforms standard ResNet (Fig.\ref{fig:ob1}(a)). The sparse expert mechanism allows the model to learn diverse data patterns within a unified parameter space, effectively enhancing representation power and convergence efficiency without incurring excessive computational costs.

% \textbf{Observation 2: Non-IID Distribution Severely Degrades MoE Performance.} In contrast, under non-IID conditions, the performance of ResNet-MoE degrades sharply, falling below the baseline (Fig.\ref{fig:ob1}(b)). 
\textbf{Observation 2: Non-IID Distribution Severely Degrades MoE Performance.} In contrast, under non-IID conditions ($\alpha=0.1$), ResNet-MoE accuracy drops to 50.60\%, significantly below ResNet's 55.78\%, a 5.18\% degradation as shown in Fig.\ref{fig:ob1}(b). 
We attribute this failure to two key factors: (1) \textit{Expert Drift}, where experts optimized on disjoint local datasets diverge semantically, leading to a fragmented global pool; and (2) \textit{Gating Inconsistency}, where independently trained routers develop conflicting selection policies, preventing coherent expert utilization during aggregation.

\textbf{Observation 3: The Generalization-Personalization Dilemma.} As illustrated in Fig.\ref{fig:ob3}, existing methods struggle to satisfy both objectives simultaneously. Global aggregation methods (e.g., FedAvg with 50.60\% global but 30.87\% personalized accuracy) sacrifice personalization for global stability, while personalized approaches (e.g., pFedMoE with 74.96\% personalized accuracy but losing global generalization capabilities) overfit to local distributions. No existing method achieves strong performance on both metrics.

% \textbf{Observation 3: The Generalization-Personalization Dilemma.} As illustrated in Fig.\ref{fig:ob3}, existing methods struggle to satisfy both objectives simultaneously. Global aggregation methods (e.g., FedAvg) sacrifice local personalization for global stability, while personalized approaches (e.g., pFedMe) often overfit local distributions, losing global generalization capabilities.

\textbf{Summary and Motivation:} These observations reveal a critical insight: \textit{the failure of MoE under non-IID stems not from the architecture itself, but from the lack of coordination mechanisms during federated aggregation}. This motivates our design of \textbf{FedCoE}, which introduces: (1) a \textit{shared semantic gating network} to synchronize routing policies across heterogeneous clients and maintain global semantic alignment, (2) a \textit{consistency-driven expert aggregation strategy} to selectively aggregate knowledge from semantically aligned clients into each global expert while redistributing personalized expert combinations back to clients for local adaptation, and (3) an \textit{adaptive cold-start mechanism} to enable instant personalized initialization for new clients with unseen distribution. FedCoE aims to resolve the generalization-personalization dilemma by constructing a unified solution that bridges global generalization, local personalization, and cold-start adaptation.

% \textbf{Summary and Motivation:} These observations reveal a critical gap: while MoE holds the potential to bridge the trade-off between generalization and personalization, its efficacy in FL is severely hampered by structural incoherence under heterogeneity. This motivates the design of \textbf{FedCoE}, a framework that explicitly addresses these challenges. By introducing a \textit{shared server-side gating mechanism} to enforce routing consistency and a \textit{correlation-aware aggregation strategy} to mitigate expert drift, we aim to unlock the full potential of MoE in federated learning. Our goal is to construct a unified model that not only generalizes well globally but also adapts robustly to local clients—even those with unseen distributions (cold-start)—thereby resolving the dilemma highlighted in Observation 3.

\begin{figure*}[t]
    \centering    \includegraphics[width=\linewidth]{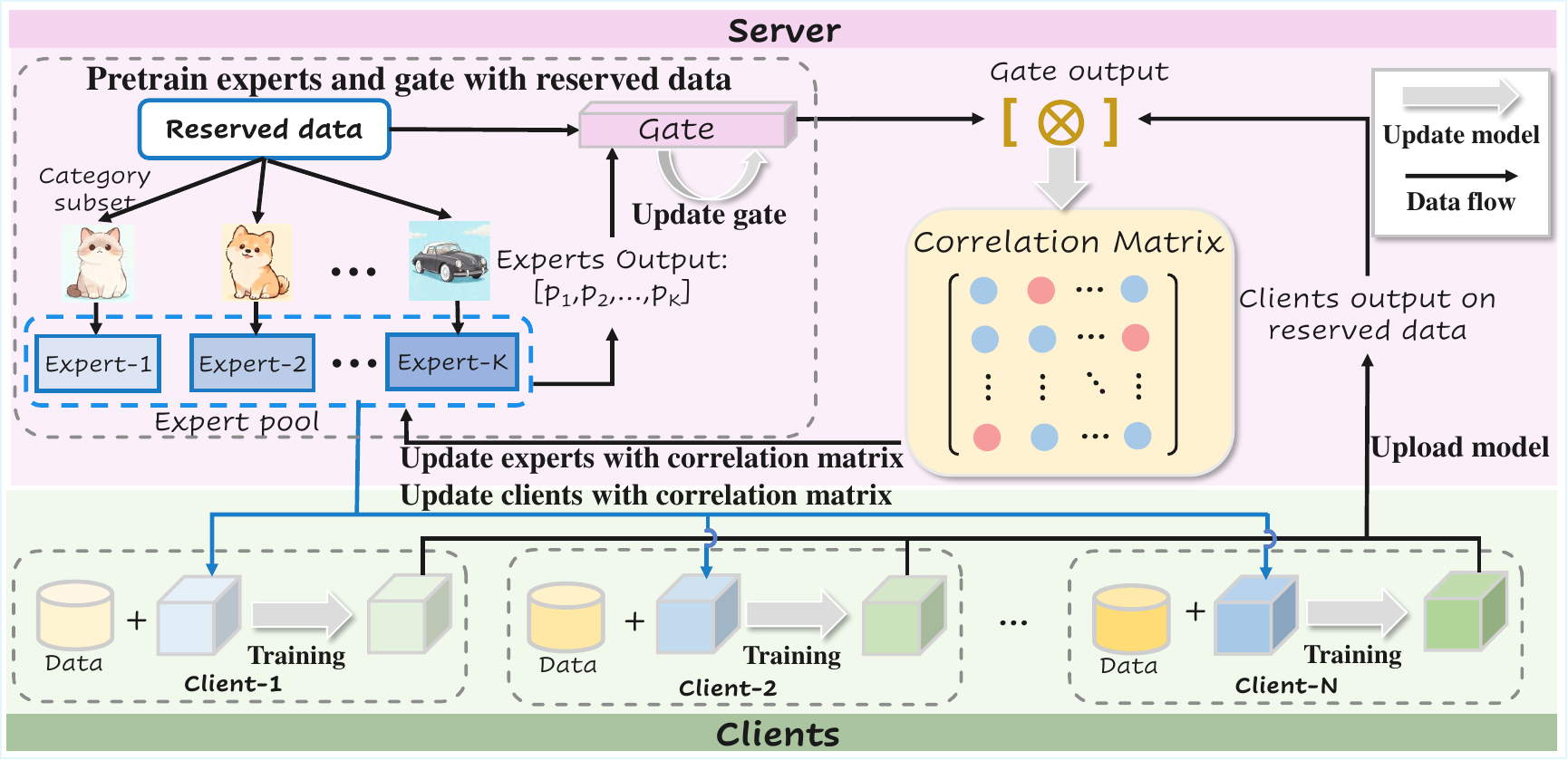}
    \caption{Overview of the proposed FedCoE framework, showing the hierarchical dual-level MoE architecture where global experts and local client models are coordinated through a shared gating network, along with the two-phase learning pipeline comprising pre-training initialization and correlation-aware federated optimization.}
    \label{fig:overview}
\end{figure*}

\section{Algorithm Design}\label{sec:alg}

In this section, we present the detailed design of FedCoE. We first provide an overview of the framework architecture, followed by the two-phase learning pipeline that initializes experts and performs federated optimization. We then introduce the shared semantic gating network for routing consistency, the consistency-driven expert aggregation strategy for expert updates, and the adaptive cold-start mechanism that enables instant personalization for new clients. To facilitate understanding, the key notations are summarized in Table~\ref{tab:notation}.

\subsection{Overview of FedCoE Framework}
\begin{figure*}[]
    \centering \includegraphics[width=1\linewidth]{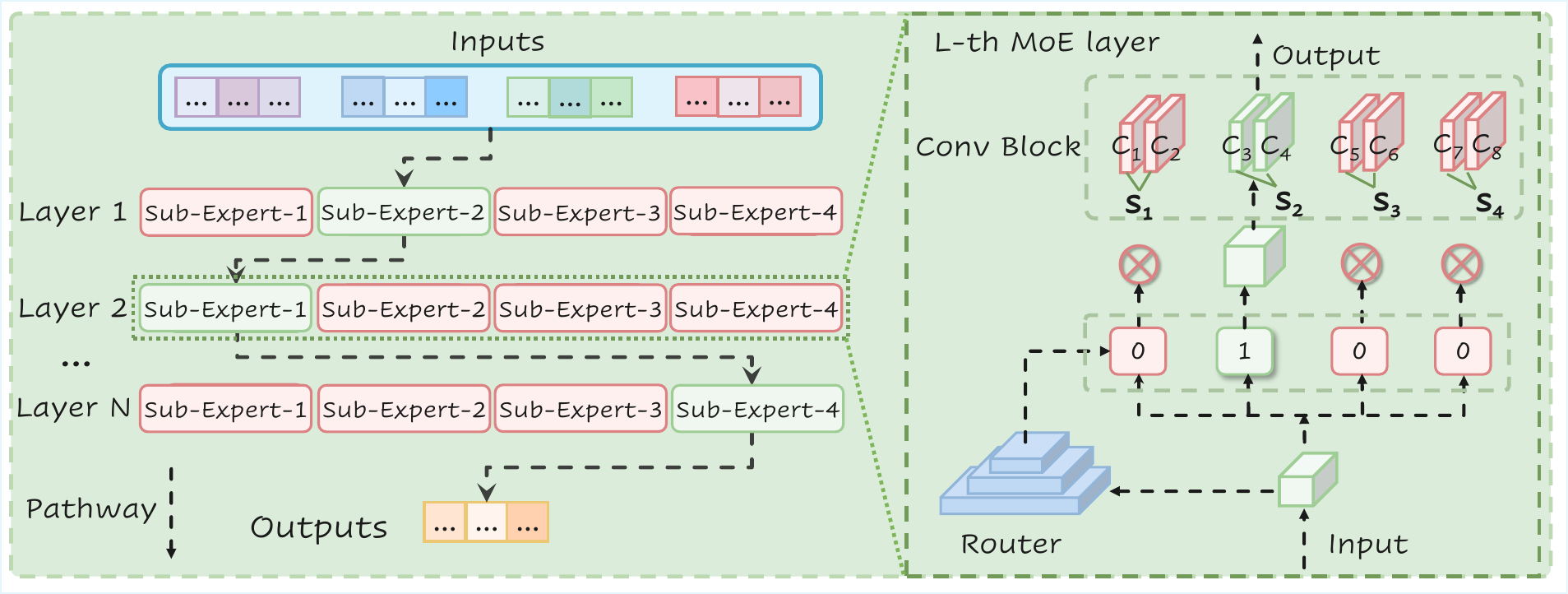}
    \caption{The detailed structure of ResNet-MoE~\cite{zhang2023robust}, containing multiple internal experts within each layer.}
    \label{fig:moe}
\end{figure*}

Building on the above analysis, we propose FedCoE, a Federated Coordinated dual-level mixture-of-Experts framework. 
As illustrated in Fig.~\ref{fig:overview}, FedCoE employs a hierarchical MoE structure where both the server and clients adopt ResNet-MoE~\cite{zhang2023robust} as the backbone architecture (see Fig.~\ref{fig:moe}). 
This design integrates multiple lightweight convolutional experts within each residual block, managed by a router that activates only the top-$k$ most relevant experts. 
While sharing this unified architectural backbone, the server and client models operate on distinct inputs to fulfill their specific roles. 
\textbf{At the client level}, the local ResNet-MoE models are trained on private local datasets, allowing their internal experts to specialize in fine-grained, personalized feature extraction.
\textbf{At the server level}, the global expert pool comprises experts that are optimized using category-specific samples from a reserved dataset, ensuring that each global expert captures distinct semantic patterns across the federation.

These two levels are coordinated through two key mechanisms: a shared gating network residing on the server that anchors the routing policy to ensure semantic consistency, and a client-expert correlation matrix that quantifies the alignment between client distributions and global experts to guide aggregation and mitigate expert drift. 
The framework operates in two complementary phases. 
In the \textbf{pre-training phase}, the server utilizes a small reserved dataset to pre-train both the global expert models and the shared gating network, establishing diverse expert specializations and an initial routing policy. 
In the \textbf{federated optimization phase}, the framework dynamically refines the relationships between clients and global experts through the correlation matrix, continuously updating the expert pool and gating network. 
This design also naturally addresses the cold-start problem: for a new client, the gating network can immediately identify and retrieve the most relevant experts, enabling high performance without extensive local training.

\begin{table}[t]
\centering
\caption{Summary of Key Notations}
\label{tab:notation}
\renewcommand{\arraystretch}{1.2}
\begin{tabular}{c|l}
\hline
\textbf{Notation} & \textbf{Description} \\
\hline
$N$ & Total number of clients ($i=1,\dots,N$) \\
$K$ & Total number of global experts ($j=1,\dots,K$) \\
$k$ & Number of active experts selected per input (i.e., top-$k$) \\
$\mathcal{D}_i$ & Local private dataset held by client $i$ \\
$\mathcal{D}_s$ & Reserved public dataset maintained by the server \\
$E_j(\cdot)$ & The $j$-th global expert network \\
$G(\cdot)$ & Shared semantic gating network \\
$s_j(x)$ & Suitability score of expert $j$ for input $x$ \\
$d$ & Dimension of the feature representation \\
$h \in \mathbb{R}^d$ & Feature representation extracted by the backbone \\
$p$ & Routing probability distribution \\
$W_g$ & Learnable weights of the gating network ($W_g \in \mathbb{R}^{K \times d}$) \\
$\mathcal{J}$ & Set of indices of the top-$k$ selected global experts \\
$\mathbf{M}$ & Client-Expert correlation matrix ($\mathbf{M} \in \mathbb{R}^{K \times N}$) \\
$a_i(x)$ & Client $i$'s confidence on the ground-truth class \\
$\boldsymbol{\theta}$ & Model parameters (e.g., $\boldsymbol{\theta}_{C_i}$ for client $i$) \\
$\mathbf{M}^{(E)}$ & Row-normalized correlation matrix for expert updates \\
$\mathbf{M}^{(C)}$ & Column-normalized correlation matrix for client updates \\
$\mathcal{S}_j$ & Set of indices of the top relevant clients for expert $j$ \\
$\tau$ & Smoothing factor controlling the model update strength \\
$N_{\text{top}}$ & Number of selected clients per expert \\
$\mathcal{S}_j$ & Set of indices of the top-$N_{\text{top}}$ relevant clients for expert $j$ \\
$\boldsymbol{\alpha}_{\text{new}}$ & Semantic profile vector for a new client ($\boldsymbol{\alpha}_{\text{new}} \in \mathbb{R}^K$) \\
\hline
\end{tabular}
\end{table}

\subsection{Two-Phase FedCoE Learning Pipeline}
FedCoE employs a two-phase learning pipeline to establish and refine the expert-client alignment. 
The pre-training phase addresses the challenge of initializing semantically meaningful experts and a consistent routing policy before federation begins. 
The subsequent federated optimization phase dynamically adapts the global expert pool and gating network based on client feedback, ensuring continuous alignment under evolving data distributions. 
We detail each phase below.

\subsubsection{Pre-training Phase}
The pretraining phase aims to establish a semantically  initialization for both the expert and gating networks before federated optimization begins. This stage ensures that experts acquire diverse and complementary feature specializations, while the gating network learns an initial routing policy that reflects expert relevance. Specifically, the server maintains a small reserved dataset $\mathcal{D}_s = \{ (x_m, y_m) \}_{m=1}^{|\mathcal{D}_s|}$ sampled from a representative global distribution. Each expert network $\{ E_j \}_{j=1}^{K}$ is pretrained independently on a disjoint subset of the reserved dataset $\mathcal{D}_s$. Specifically, $\mathcal{D}_s$ is partitioned into class-wise subsets $\{\mathcal{D}_s^{j}\}_{j=1}^{K}$, and each expert $ E_j $ is trained on $\mathcal{D}_s^{j}$ to specialize in distinct semantic regions of the data distribution.

After the experts are pretrained, the server trains the gating network $G(\cdot)$ to learn the mapping between inputs and the most suitable experts. Unlike conventional supervised training, the gating network is not optimized using ground-truth labels but rather through soft supervision derived from the pretrained experts. For an input sample $x$, each expert $ E_j $ produces a prediction with confidence $ p_j(x)$. These confidence values are normalized to form an expert suitability distribution:
\begin{equation}
s_j(x) = \frac{\exp(p_j(x))}{\sum_{l=1}^{K} \exp(p_l(x))},
\label{eq:suitability}
\end{equation}
where $s_j(x)$ represents the relative confidence of expert $E_j$ for the given input.

The gating network outputs a probability vector $G(x) = [G_1(x), \dots, G_K(x)]$, which is trained to align with the expert suitability distribution via Kullback–Leibler divergence:
\begin{equation}
\mathcal{L}_{\text{gate}} = \text{KL}\big(s(x) \parallel G(x)\big)
= \sum_{j=1}^{K} s_j(x) \log \frac{s_j(x)}{G_j(x)}.
\label{eq:kl_loss}
\end{equation}

After pretraining, the expert networks provide a diverse and well-structured knowledge base, while the gating network learns an initial semantic alignment between inputs and experts.

\subsubsection{Federated Optimization Phase}
In the federated optimization stage, each communication round proceeds as follows: 
a) Distribution: Clients receive a personalized combination of global expert models from the server based on the routing policy and perform local training on private data to update their internal experts. 
b) Uploading: After local updates, clients upload their model parameters to the server. 
c) Aggregation: The server constructs a client-expert correlation matrix based on the gating outputs and prediction distributions. This matrix quantitatively measures the semantic relevance between each client's local distribution and the global experts. 
d) Updating: Using this correlation matrix, the server updates the global expert Pool through correlation-weighted aggregation and refines the shared gating network to maintain routing consistency across clients.

\subsection{Shared Semantic Gating Network}
The shared semantic gating network is the central coordination mechanism in FedCoE, serving to unify expert selection policies across heterogeneous clients and eliminate gating inconsistency. 
Unlike local routers that manage fine-grained internal experts within each client's network, this server-side gating network operates at a higher level to coordinate global expert models. 
It fulfills three critical functions: modeling client-expert correlations during federated optimization, enabling zero-shot expert assembly for cold-start clients, and supporting global model inference. 
We elaborate on each function below.

Formally, let $h \in \mathbb{R}^d$ denote the feature representation of an input sample extracted by the backbone network. The gating network $G(\cdot)$, parameterized by weights $W_g \in \mathbb{R}^{K \times d}$, computes the routing probability distribution $p \in \mathbb{R}^K$ as follows:
\begin{equation}
p = G(h) = \text{Softmax}(W_g \cdot h),
\end{equation}
where each element $p_j$ represents the confidence score of assigning the input to the $j$-th global expert.

In our framework, this gating network serves two distinct functions depending on the operational phase:

\subsubsection{Correlation Modeling during Federated Optimization (Server-Side)}During the standard federated training loop, the gating network resides exclusively on the server. It is utilized to construct the \textit{client-expert correlation matrix}. By evaluating the semantic alignment between uploaded client model updates and the global experts, the gating network guides the server in generating personalized aggregation weights. This centralized design ensures that expert selection criteria remain consistent, preventing the divergence often seen when clients train independent routers.

\subsubsection{Expert Initialization for Cold-Start Clients (Client-Side Deployment) }The only scenario where the gating network is deployed to the client side is during the cold-start phase. When a new client joins the federation with no prior model history, the server transmits the pre-trained gating network to this client. The new client uses this network to perform a one-time inference on its local data to identify the top-$k$ most relevant experts. This mechanism allows the new client to immediately retrieve a highly suitable subset of experts from the global pool, thereby bypassing the slow convergence typically associated with random initialization.

\subsubsection{Global Model Inference}
Beyond its role in coordination, the gating network and the global expert pool collectively constitute the global MoE model. During the inference or testing phase on the server, the gating network dynamically computes routing weights for incoming test samples. The final prediction is generated by the weighted combination of outputs from the selected top-$k$ global experts:
\begin{equation}
y = \sum_{j \in \mathcal{J}} p_j \cdot E_j(h),
\end{equation}
where $\mathcal{J}$ denotes the set of indices corresponding to the top-$k$ probability values in $p$. This mechanism allows the server to evaluate global generalization performance directly.

\begin{figure}[]
    \centering
    \includegraphics[width=1\linewidth]{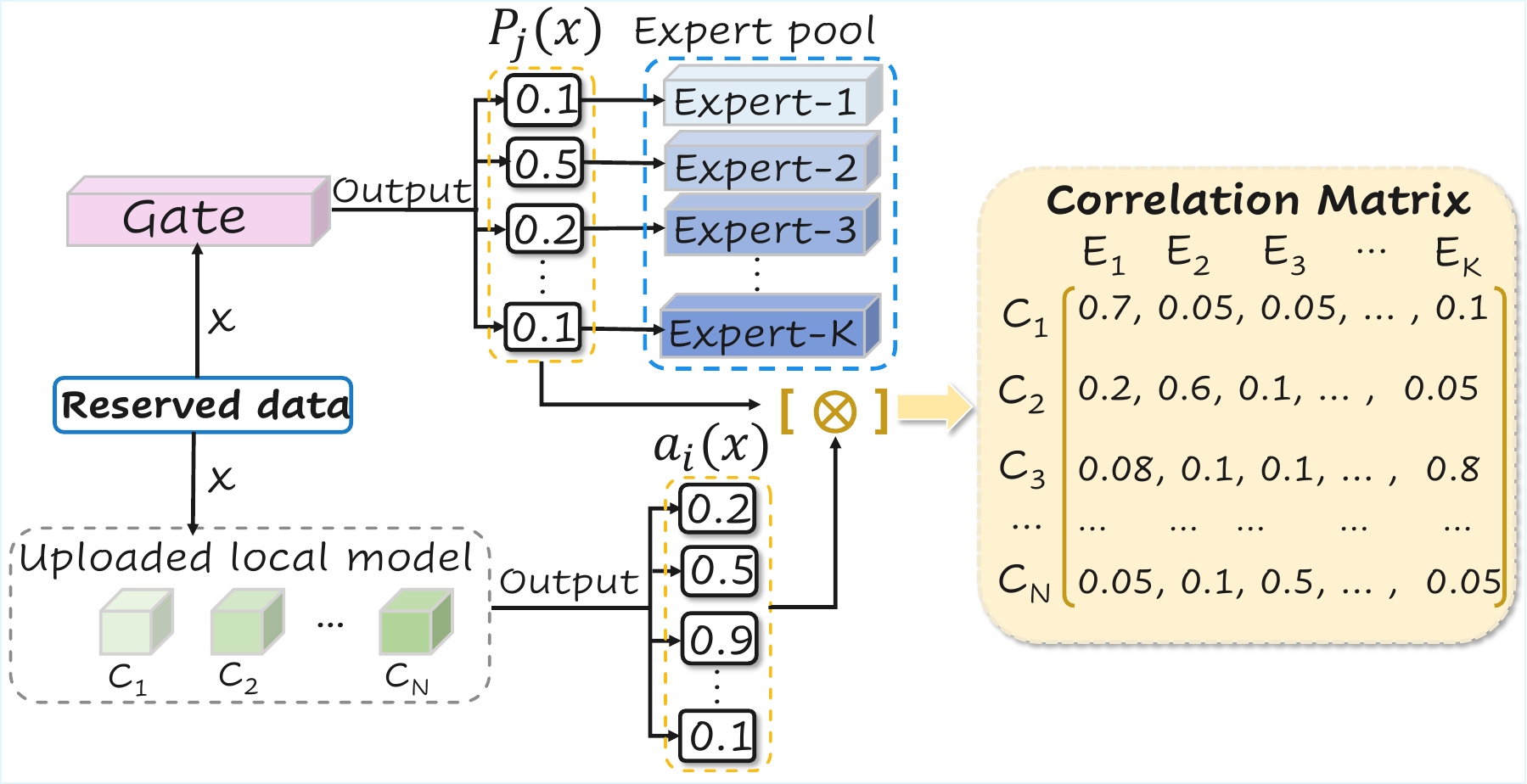}
    \caption{Process of Constructing the Correlation Matrix}
    \label{fig:corr-matrix}
\end{figure}

\subsection{Consistency-driven Expert Aggregation Strategy}
The consistency-driven expert aggregation strategy is designed to mitigate expert drift, a phenomenon where experts trained on disjoint client distributions diverge semantically and lose global coherence. 
By leveraging the gating network outputs, we construct a client-expert correlation matrix that quantifies semantic relevance, 
    enabling selective aggregation where each expert absorbs knowledge only from semantically aligned clients. 
This subsection details the correlation matrix construction and the selective update mechanism.

\subsubsection{Correlation Matrix Construction}
To model the semantic relevance between experts and clients, FedCoE constructs a correlation matrix that quantifies how each expert contributes to each client’s prediction behavior.
As shown in Fig.\ref{fig:corr-matrix}, this matrix serves as the foundation for correlation-weighted aggregation and personalized expert redistribution in the subsequent optimization phase.

Concretely, the server evaluates all client models $\{C_i\}_{i=1}^{N}$ and the global gating network $G(\cdot)$ on the reserved dataset $\mathcal{D}_s$. For each input $x \in \mathcal{D}_s$, the prediction confidence of client $i$ on the ground-truth class is denoted as:
\begin{equation}
a_i(x) = P(y|x; \theta_{C_i}),
\label{eq:client_confidence}
\end{equation}

and the gating probability of expert $j$ for the same input is given by the gating output defined in Sec.~\ref{sec:alg}-C:
\begin{equation}
p_j(x) = [G(x)]_j,
\label{eq:gate_weight}
\end{equation}
where $p_j(x)$ corresponds to the $j$-th element of the probability vector $p$.

The pairwise contribution between expert $j$ and client $i$ is measured by the outer product of their respective gating weights and confidence scores over the dataset:
\begin{equation}
\mathbf{M} = \sum_{x \in \mathcal{D}_s} p(x) \, a(x)^{\top},
\label{eq:correlation_matrix}
\end{equation}
where $p(x) \in \mathbb{R}^K$ is the gating probability vector, $a(x) = [a_1(x), \dots, a_N(x)]^\top \in \mathbb{R}^N$ is the vector of client confidences, and $\mathbf{M} \in \mathbb{R}^{K \times N}$ encodes the semantic correlation between $K$ experts and $N$ clients.

To normalize the interaction strength and avoid scale bias, FedCoE derives two normalized correlation matrices:
\begin{equation}
\mathbf{M}^{(E)} = 
\frac{\mathbf{M}}
{\mathbf{M} \mathbf{1}_N + \varepsilon}, \qquad
\mathbf{M}^{(C)} = 
\frac{\mathbf{M}}
{\mathbf{1}_K^\top \mathbf{M} + \varepsilon},
\label{eq:normalized_corr}
\end{equation}
where $\mathbf{1}_N$ and $\mathbf{1}_K$ are all-one vectors of size $N$ and $K$, respectively. $\mathbf{M}^{(E)}$ and $\mathbf{M}^{(C)}$ denote the row-normalized and column-normalized matrices used for expert aggregation and client redistribution, respectively. This correlation modeling aligns experts with semantically similar clients, stabilizing aggregation and improving personalization.

\begin{figure}[]
    \centering
    \includegraphics[width=1\linewidth]{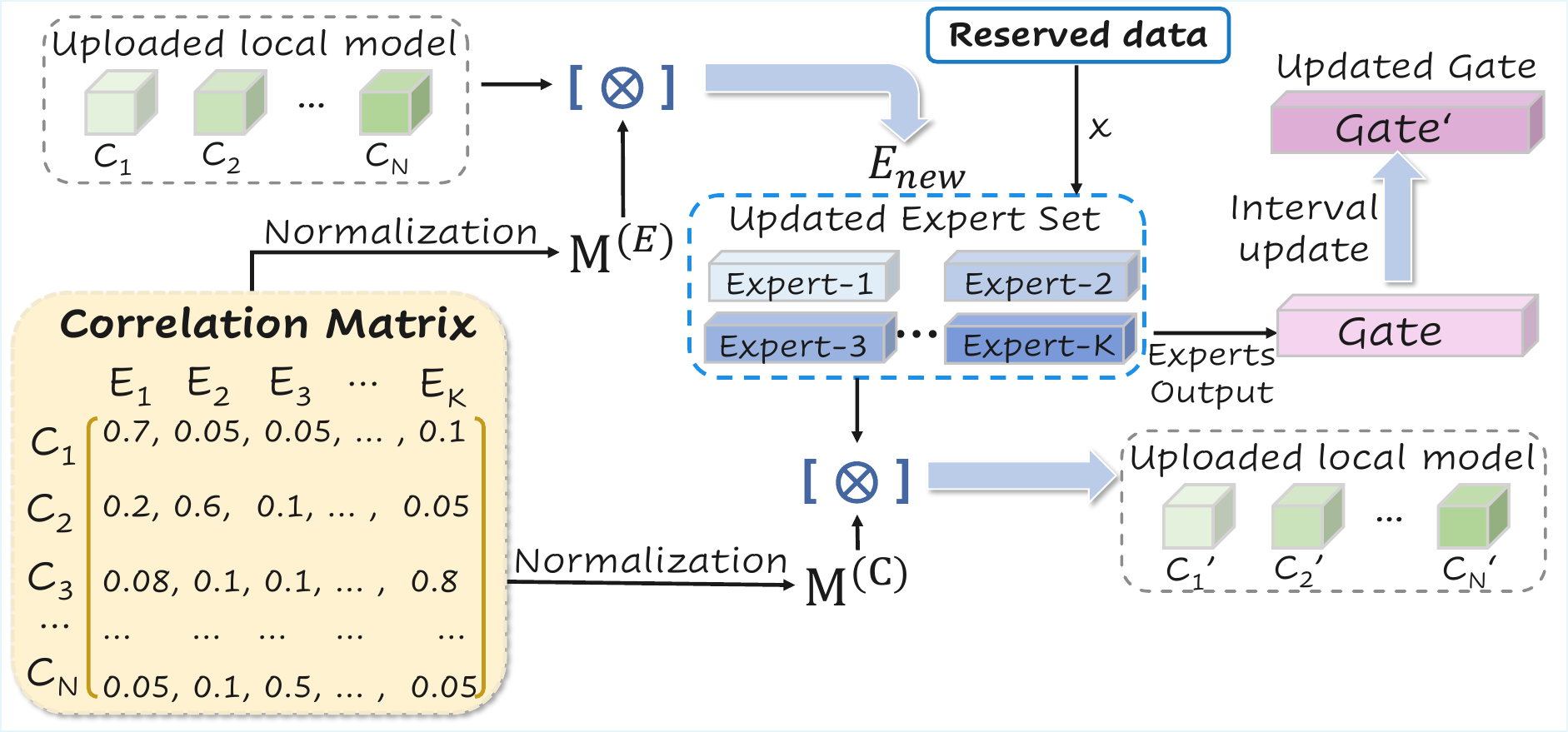}
    \caption{Process of Updating Local Model and Expert Set with Correlation Matrix}
    \label{fig:weighted_update}
\end{figure}

\subsubsection{Selective Correlation-Guided Update}
Building upon the constructed correlation matrices, the server performs a selective, consistency-driven update of the expert pool and client models in each communication round. During each communication round, as shown in Fig.\ref{fig:weighted_update}, the correlation matrix acts as a weighting mechanism that directs the joint optimization of the experts, gating network, and client models via correlation-aware aggregation and redistribution.

Specifically, the server utilizes the expert update matrix $\mathbf{M}^{(E)}$, whose entry $M_{j,i}^{(E)}$ measures the semantic correlation between expert $E_j$ and client $C_i$.
To prevent the ''dilution'' of expert knowledge by irrelevant or noisy client updates, each expert $E_j$ only aggregates a subset of the most relevant clients $\mathcal{S}_j$. The update rule is formulated as:
\begin{equation}
\boldsymbol{\theta}_{E_j}^{(t+1)} 
= (1 - \tau)\, \boldsymbol{\theta}_{E_j}^{(t)} 
+ \tau \sum_{i \in \mathcal{S}_j} \bar{M}_{j,i}^{(E)} \, \boldsymbol{\theta}_{C_i}^{(t)},
\label{eq:expert_update}
\end{equation}
where $\tau \in [0,1]$ is a smoothing factor. $\bar{M}_{j,i}^{(E)}$ represents the locally re-normalized weight derived to ensure the aggregation scale remains stable within the selected subset:
\begin{equation}
\bar{M}_{j,i}^{(E)} = \frac{M_{j,i}^{(E)}}{\sum_{n \in \mathcal{S}_j} M_{j,n}^{(E)}}.
\end{equation}
This selective mechanism allows each expert to exclusively absorb knowledge from semantically aligned clients, thereby reinforcing its domain specialization while filtering out interference from disparate distributions.

To adapt to the evolving expert representations, the server periodically fine-tunes the gating network on the reserved dataset. This fine-tuning step recalibrates the routing policy so that the gating outputs remain aligned with the updated expert semantics.

After server-side aggregation, each client model is refreshed using the client update weighted matrix $\mathbf{M}^{(C)}$, which distributes the updated experts back to clients in a personalized manner:
\begin{equation}
\boldsymbol{\theta}_{C_i}^{(t+1)} 
= (1 - \tau)\, \boldsymbol{\theta}_{C_i}^{(t)} 
+ \tau \sum_{j=1}^{K} M_{j,i}^{(C)} \, \boldsymbol{\theta}_{E_j}^{(t)}.
\end{equation}
This allows each client $i$ to flexibly combine knowledge from multiple relevant experts ($j=1 \dots K$) to serve its local data distribution.

\begin{figure}[]
    \centering
    \includegraphics[width=1\linewidth]{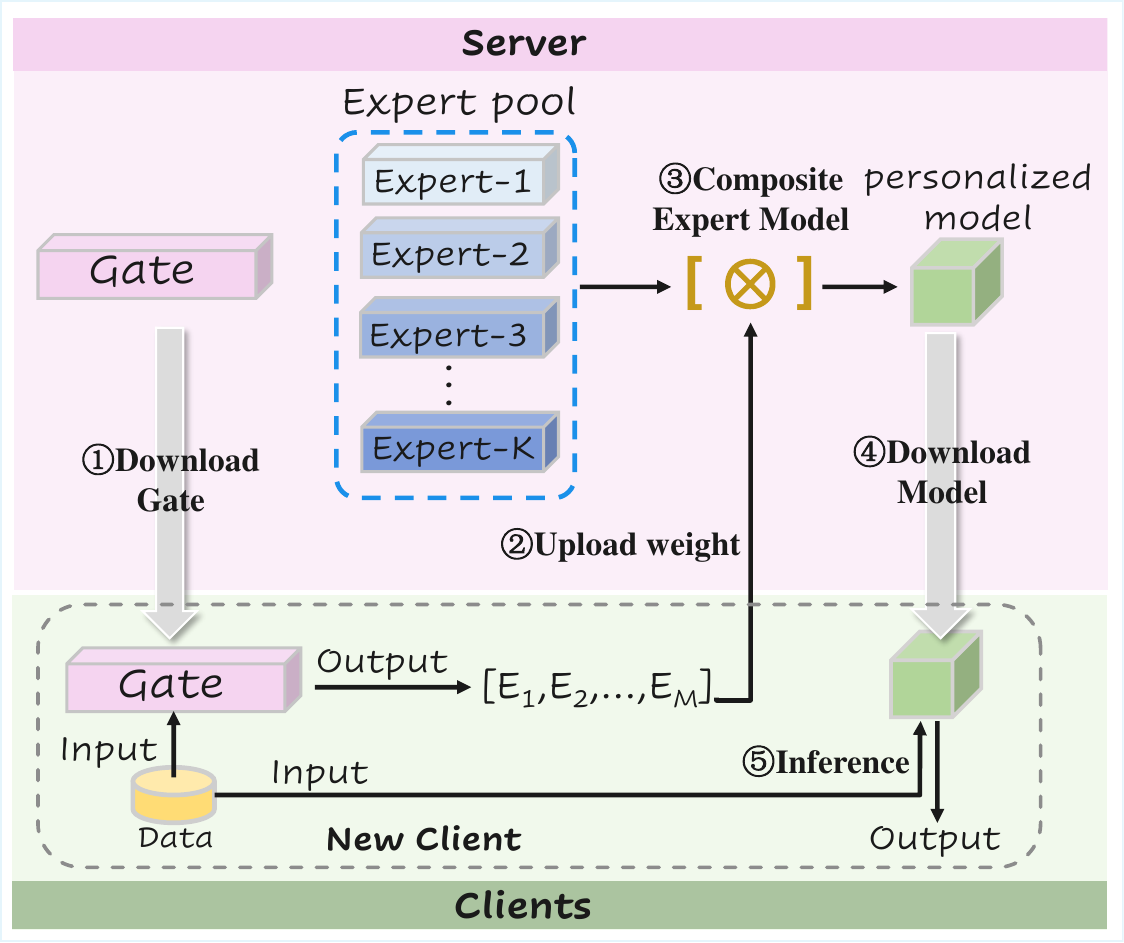}
    \caption{The workflow of the adaptive cold-start mechanism for new clients corresponds to the algorithmic steps detailed in Section \ref{sec:cold_start}: 
    \textcircled{1} Downloading the shared gating network; 
    \textcircled{2} Uploading the local semantic profile derived from the gating network; 
    \textcircled{3} Server-side expert assembly and \textcircled{4} Distribution of the personalized model; 
    \textcircled{5} Immediate local inference without gradient updates.}
    \label{fig:cold_start}
\end{figure}

\subsection{Adaptive Cold-Start Mechanism}
\label{sec:cold_start}

The adaptive cold-start mechanism enables new clients to obtain high-performance personalized models immediately upon joining the federation, without any local gradient updates. 
It addresses a critical limitation of existing methods that require extensive fine-tuning under data scarcity. 
The core idea is to leverage the shared gating network as a semantic probe: new clients use it to profile their local data distribution and retrieve a tailored expert combination from the global pool. 
As illustrated in Fig.~\ref{fig:cold_start}, the mechanism operates in four steps: gating network deployment, local semantic profiling, profile-guided expert assembly, and immediate inference.

\subsubsection{Gating Network Deployment}
When a new client $i_{\text{new}}$ joins the federation, instead of downloading the massive full model or a randomly averaged global model, it requests only the lightweight shared gating network $G(\cdot)$ from the server. This minimizes the initial communication overhead.

\subsubsection{Local Semantic Profiling}
The new client utilizes the downloaded gating network to evaluate its local data distribution $\mathcal{D}_{\text{new}}$. Although $\mathcal{D}_{\text{new}}$ may be small, it contains sufficient semantic information to identify task relevance. The client performs a forward pass on its local samples using $G(\cdot)$ to compute the routing probability for each sample. These probabilities are then averaged to form a global semantic profile vector $\boldsymbol{\alpha}_{\text{new}} \in \mathbb{R}^K$:
\begin{equation}
    \boldsymbol{\alpha}_{\text{new}} = \frac{1}{|\mathcal{D}_{\text{new}}|} \sum_{x \in \mathcal{D}_{\text{new}}} G(x),
    \label{eq:profiling}
\end{equation}
where the $j$-th element of $\boldsymbol{\alpha}_{\text{new}}$, denoted as $\alpha_{\text{new}, j}$, quantifies the relevance of the $j$-th global expert to the new client's local task. This profile effectively maps the unknown local distribution into the known semantic space of the global experts.

\subsubsection{Profile Upload and Expert Assembly}
The client uploads the low-dimensional profile vector $\boldsymbol{\alpha}_{\text{new}}$ to the server. Importantly, no raw data or gradient information is transmitted, preserving privacy.
Upon receiving $\boldsymbol{\alpha}_{\text{new}}$, the server synthesizes a personalized initialization model $\boldsymbol{\theta}_{\text{init}}$ by aggregating the parameters of global experts $\{E_j\}_{j=1}^K$ according to the profile weights:
\begin{equation}
    \boldsymbol{\theta}_{\text{init}} = \sum_{j=1}^{K} \alpha_{\text{new}, j} \cdot \boldsymbol{\theta}_{E_j}.
    \label{eq:assembly}
\end{equation}
This aggregated model is then transmitted back to the client.

\subsubsection{Immediate Inference and Integration}
Upon receiving $\theta_{init}$, the new client loads these parameters into its local model structure. Since $\theta_{init}$ is constructed from well-trained global experts specifically matched to the client's semantic profile, the client can immediately perform inference on local test data with competitive accuracy, even before any local gradient updates occur. 
This capability enables zero-shot or few-shot adaptation, allowing the client to serve online requests instantly. Furthermore, should the client participate in subsequent federated training rounds, this personalized model serves as a superior initialization point (``warm start"), significantly accelerating convergence compared to random or average initialization.

Through this mechanism, the new client instantly acquires a high-performance model composed of the most relevant capabilities from the global expert pool. Whether the client's distribution is similar to existing ones or completely unique, this weighted combination provides a robust starting point without requiring iterative training, effectively solving the cold-start challenge.

\section{Performance Evaluation}\label{sec:performance_evaluation}
\subsection{Experimental Setup}

\begin{table*}[t]
\centering
\setlength{\tabcolsep}{2.8mm} %2.2mm
\renewcommand{\arraystretch}{1}
\sethlcolor{color_pink}
\caption{Comparison of Global and Personalized Accuracy (\%) Across Different Data Distributions.}

\label{tab1:noniid}
% \adjustbox{width=1\linewidth}{
{
\small
\begin{tabular}{lcccccccc}
    \toprule
    % --- 这是新的三行表头 ---
    
    % --- Row 1 ---
    \multirow{3}{*}{\textbf{Methods}} % Col 1: 跨3行
    & \multicolumn{2}{c}{\multirow{2}{*}{\textbf{IID}}} % Col 2-3: 跨2行, 跨2列
    & \multicolumn{6}{c}{\textbf{Dirichlet Distribution}} % Col 5-12: 跨8列
     \\% Col 13 (spacer)
    
    % Rules after Row 1
     \cmidrule{4-9}
    
    % --- Row 2 (已修正) ---
    & % Col 1 (Methods 的占位符)
    \multicolumn{2}{c}{} % Col 2-3 (IID 的 \multirow 占位符，必须合并)
    & \multicolumn{2}{c}{\textbf{$\alpha$ = 1}} % Col 5-6
    & \multicolumn{2}{c}{\textbf{$\alpha$ = 0.5}} % Col 8-9
    & \multicolumn{2}{c}{\textbf{$\alpha$ = 0.1}} % Col 11-12
     \\ % Col 13 (spacer)
    
    % Rules after Row 2 (under alphas)
    \cmidrule{2-3}\cmidrule{4-5} \cmidrule{6-7} \cmidrule{8-9}

    % --- Row 3 ---
    & \textbf{$G_{acc}$} & \textbf{$P_{acc}$} % Col 2-3 (for iid)
    & \textbf{$G_{acc}$} & \textbf{$P_{acc}$} % Col 5-6 (for a=0.1)
    & \textbf{$G_{acc}$} & \textbf{$P_{acc}$} % Col 8-9 (for a=0.5)
    & \textbf{$G_{acc}$} & \textbf{$P_{acc}$} % Col 11-12 (for a=1)
     \\ % Col 13 (spacer)
    % --- 表头结束 ---
%%%%%%%%%%%%%% 
\midrule[0.5pt]      
FedAvg\cite{mcmahan2017communication}&82.34 &81.51&\cellcolor{color_blue}81.26 &67.72&75.69 &81.81 &50.60 &30.87\\
FedProx\cite{li2020federated}&\cellcolor{color_blue}84.11 &\cellcolor{color_blue}80.76&78.34 &76.89& \cellcolor{color_blue}80.05 &\cellcolor{color_blue}86.52 &29.17 &87.52\\
SCAFFOLD\cite{karimireddy2020scaffold}&61.80 &58.36&73.68 &71.29&44.41 &35.84 &33.40 &23.48\\
FedMoEKD\cite{liang2025mixture}&-- &42.40&-- &45.00&-- &42.40 &-- &77.67\\
pFedMoE\cite{yi2024pfedmoe}&-- &22.13&-- &37.54&-- &50.78 &-- &74.96\\
PM-MoE\cite{feng2025pm}&79.67 &80.40&74.57 &\cellcolor{color_blue}80.92&68.71 &82.34 &\cellcolor{color_pink}66.38 &\cellcolor{color_blue}90.10\\
\textbf{FedCoE (Ours)}&\cellcolor{color_pink}85.30 &\cellcolor{color_pink}84.48
&\cellcolor{color_pink}85.10 &\cellcolor{color_pink}85.86 &\cellcolor{color_pink}85.55 &\cellcolor{color_pink}90.26 &\cellcolor{color_blue}63.35 &\cellcolor{color_pink}91.84\\
 \bottomrule
       \end{tabular}
% }
}
\begin{tablenotes}
\vspace{\smallskipamount}
\footnotesize
\item \parbox{\linewidth}{ \textit{Note: The symbol “-” indicates that $G_{acc}$ is not evaluated for methods without a unified global model. $G_{acc}$ denotes the accuracy of the aggregated global model, and $P_{acc}$ represents the average personalized accuracy across clients. The \hl{best}/\sethlcolor{color_blue}\hl{second-best} are color coded.}}
\end{tablenotes}
   
\end{table*}

\subsubsection{Dataset}
We evaluate our framework on three image classification benchmarks: CIFAR-10~\cite{krizhevsky2009learning}, CIFAR-100~\cite{krizhevsky2009learning}, and Tiny-ImageNet~\cite{deng2009imagenet}. 
CIFAR-10 serves as the primary benchmark, which consists of 60,000 $32 \times 32$ color images distributed across 10 distinct classes. The dataset is explicitly split into 50,000 training images and 10,000 test images. 
To evaluate scalability under increased complexity, CIFAR-100 comprises 100 classes, also following the standard split of 50,000 training and 10,000 test images. Tiny-ImageNet further increases the difficulty with 200 classes and a higher resolution of $64 \times 64$, containing 100,000 training images and 10,000 validation images used for testing.
For the server-side reserved dataset $\mathcal{D}_s$, we randomly sample 20\% of the global test set to form a representative subset. To support the disjoint expert initialization, $\mathcal{D}_s$ is strictly partitioned into class-wise subsets $\{\mathcal{D}_s^{j}\}_{j=1}^{K}$, where each expert $E_j$ is pre-trained on a corresponding subset $\mathcal{D}_s^{j}$ to acquire distinct semantic specializations.
% \subsubsection{Data set}:We comprehensively evaluate the efficacy of the proposed FedCoE framework on the CIFAR-10 dataset, a widely adopted benchmark in federated learning research. Its inherent class diversity provides a rigorous testing ground for simulating varying degrees of non-IID conditions. Our experimental results demonstrate that FedCoE not only consistently outperforms state-of-the-art baselines across heterogeneous settings but also maintains stable convergence, exhibiting superior robustness against statistical distribution shifts.

\subsubsection{Federated Learning Configuration}
To rigorously assess performance under statistical heterogeneity, we simulate non-IID data distributions using Dirichlet sampling. We partition the CIFAR-10 training data among $N=10$ clients by sampling label distributions from a Dirichlet distribution $Dir(\alpha)$. The concentration parameter $\alpha$ controls the degree of heterogeneity: a smaller $\alpha$ (e.g., 0.1) induces extreme non-IID partitions where clients possess data from only a few classes, whereas a larger $\alpha$ (e.g., 1.0) approaches an IID distribution.
For evaluation, we reserve 20\% of each client's local data as a hold-out test set to measure personalized performance (Personalized Accuracy), while the remaining 80\% constitutes the training set. 
Global Accuracy is evaluated on remaining 80\% of the global test set (excluding $\mathcal{D}_s$).

\subsubsection{Model Architecture}
We adopt ResNet-MoE \cite{zhang2023robust} as the backbone architecture for all experiments, specifically using a ResNet-18 variant with $K=2$ experts per residual block by default. This model embeds multiple lightweight convolutional expert modules within the residual blocks. It employs a two-level gating mechanism: an internal router within each residual block dynamically selects experts based on sample features, while our proposed shared gating network, a hybrid CNN-MLP module, manages the high-level routing to the global expert pool.

\subsubsection{Baseline Methods for Comparison}
To comprehensively evaluate the effectiveness of FedCoE, we compare it with representative federated learning baselines categorized as follows:

\begin{itemize}
\item[-] \textbf{Global Aggregation Methods}: a) FedAvg \cite{mcmahan2017communication} aggregates local model parameters by simple averaging to build a shared global model, serving as a fundamental benchmark for global model performance. b) FedProx \cite{li2020federated} introduces a proximal regularization term in local updates to mitigate model divergence under heterogeneous data. c) SCAFFOLD \cite{karimireddy2020scaffold} employs control variates to correct local update directions, reducing client drift caused by non-IID distributions.

\item[-] \textbf{Personalized Federated Method:}  FedMoEKD \cite{liang2025mixture} adopts a dual-expert architecture, where global and local experts are collaboratively optimized through knowledge distillation.

\item[-] \textbf{MoE-based Federated Methods:} a) pFedMoE \cite{yi2024pfedmoe} decomposes models into shared and private experts, enabling adaptive expert selection via client-specific gating networks. b) PM-MoE \cite{feng2025pm} aggregates client-specific parameters through a learnable gating network that dynamically weights expert contributions from a global parameter pool.

\end{itemize}

\subsubsection{Implementation Details}
All experiments are implemented using PyTorch~\cite{paszke2019pytorch} (version 2.0.1) on a server equipped with an Intel Xeon Gold 6330 CPU @ 2.00GHz, 512GB RAM, and four NVIDIA GeForce RTX 4090 GPUs (24GB VRAM each).
Before federated training, we perform a warm-up phase where both the global expert models and the shared gating network are pre-trained on a small reserved public dataset for 30 epochs to initialize semantic representations. The federated training proceeds for 1,000 communication rounds. To ensure the routing policy remains current, the shared gating network is updated every 10 rounds involving the server. For local optimization, we use Stochastic Gradient Descent (SGD) with a learning rate of 0.1, momentum of 0.9, and weight decay of $5 \times 10^{-4}$.

\subsection{Simulation Results}
Table \ref{tab1:noniid} summarizes the comparative performance of FedCoE against state-of-the-art baselines under varying degrees of data heterogeneity. Overall, FedCoE consistently demonstrates superior efficacy across all experimental settings, validating its robustness and adaptability to diverse data distributions.

\subsubsection{Performance across Non-IID Settings}
In scenarios characterized by IID or moderate heterogeneity ($\alpha \in \{1.0, 0.5\}$), FedCoE achieves the highest accuracy in both global and personalized evaluations. Under severe heterogeneity ($\alpha = 0.1$), while PM-MoE exhibits a slight edge in global accuracy, FedCoE retains the highest personalized accuracy.
These findings reveal a distinct performance dichotomy: global aggregation methods (e.g., FedAvg, FedProx) favor low-heterogeneity regimes, whereas personalized methods (e.g., PM-MoE, pFedMoE) excel in high-heterogeneity settings. FedCoE, however, effectively bridges this gap, maintaining competitive performance across the full spectrum of distribution shifts and achieving an optimal balance between global generalization and local personalization.

\subsubsection{Comparison with ResNet Baselines}
To further validate the effectiveness of FedCoE, we compared its performance on ResNet series models under varying levels of data heterogeneity. As shown in Table \ref{tab1:resnet}, models trained with global aggregation methods (e.g., FedAvg, FedProx) exhibit unstable performance when combined with the ResNet-MoE backbone, showing only marginal improvements under mild heterogeneity but suffering sharp degradation at high heterogeneity ($\alpha = 0.1$) due to expert drift and gating inconsistency. In contrast, while personalized federated methods (e.g., PM-MoE) achieve moderate improvements on ResNet-MoE, they still fail to fully exploit the structure’s potential. Moreover, FedCoE consistently improves the personalized accuracy on both backbones, suggesting that the proposed correlation-weighted aggregation effectively mitigates client drift. The improvement is particularly pronounced on the MoE backbone, where FedCoE not only restores model stability under highly non-IID conditions but also maintains steady gains across milder settings. These results confirm that FedCoE functions as a general aggregation enhancement mechanism while simultaneously stabilizing and strengthening MoE training, thereby unlocking the full potential of modular expert architectures in federated learning.

\begin{table*}[t]
\centering
\setlength{\tabcolsep}{2.2mm} %2.2mm
\renewcommand{\arraystretch}{1}
\sethlcolor{color_pink}
\caption{Comparison of Resnet and Resnet-MoE Performance (\%) Across Different Data Distributions.}
\label{tab1:resnet}
% \adjustbox{width=1\linewidth}{
{
\small
\begin{tabular}{lcccccc}
    \toprule
    % --- 这是新的三行表头 ---
    
    % --- Row 1 ---
    \multirow{2}{*}{\textbf{Methods}} % Col 1: 跨3行    
    & \multicolumn{2}{c}{\textbf{$\alpha$ = 1}} % Col 5-6
    & \multicolumn{2}{c}{\textbf{$\alpha$ = 0.5}} % Col 8-9
    & \multicolumn{2}{c}{\textbf{$\alpha$ = 0.1}} % Col 11-12
    \\
    % Rules after Row 1
     \cmidrule{2-7}

    % --- Row 2 ---
    & \textbf{Ori/MoE} & \textbf{$\Delta$} % Col 2-3 (for iid)
    & \textbf{Ori/MoE} & \textbf{$\Delta$} % Col 5-6 (for a=0.1)
    & \textbf{Ori/MoE} & \textbf{$\Delta$} % Col 8-9 (for a=0.5)
     \\ % Col 13 (spacer)
    % --- 表头结束 ---
%%%%%%%%%%%%%% 
\midrule[0.5pt]      
FedAvg\cite{mcmahan2017communication}&(76.19 / 67.72)&-8.47 &(80.67 / 81.81)&+1.14 &(\sethlcolor{color_blue}\hl{88.92} / 30.87)&-58.05  \\
FedProx\cite{li2020federated}&(\sethlcolor{color_blue}\hl{76.70} / 76.89)&+0.19&(\sethlcolor{color_pink}\hl{82.67} / \sethlcolor{color_blue}\hl{86.52})&+3.85 &(88.70 / 87.52)&-1.18 \\
SCAFFOLD\cite{karimireddy2020scaffold}&(73.47 / 71.29)&-2.18&(74.69 / 35.84)&-38.85&(85.39 / 23.48)&-61.91  \\
FedMoEKD\cite{liang2025mixture}&(39.40 / 45.00)&+5.6&
(50.93 / 42.40)&-8.53&(74.83 / 77.67)&+2.84  \\
pFedMoE\cite{yi2024pfedmoe}&(40.79 / 37.54)&-3.25&(50.97 / 50.78)&-0.19&(72.76 / 74.96)&+2.2  \\
PM-MoE\cite{feng2025pm} &(73.29 / \sethlcolor{color_blue}\hl{80.92})&+7.63&(71.05 / 82.34)&+11.29&(84.88 / \sethlcolor{color_blue}\hl{90.10})&+5.22 \\
\textbf{FedCoE (Ours)}&(\sethlcolor{color_pink}\hl{79.02} / \hl{85.86})&+6.84
&(\sethlcolor{color_blue}\hl{81.20} / \sethlcolor{color_pink}\hl{90.26})&+6.41 &(\sethlcolor{color_pink}\hl{90.53} / \hl{91.84})&+1.13  \\
 \bottomrule
       \end{tabular}
% }
}
\begin{tablenotes}
\vspace{\smallskipamount}
\footnotesize
\item \parbox{\linewidth}{ \textit{Note:Ori/MoE reports “ResNet / ResNet-MoE” accuracy. The $\Delta$ denotes the improvement of ResNet-MoE over ResNet across different $\alpha$ values.The \hl{best}/\sethlcolor{color_blue}\hl{second-best} are color coded.}}
\end{tablenotes}
   
\end{table*}

\subsubsection{Performance on Different Datasets}
To evaluate the scalability and generalization capability of FedCoE across varying levels of data complexity, we extended our experiments to include three benchmarks: CIFAR-10, CIFAR-100, and Tiny-ImageNet.While CIFAR-10 serves as a standard baseline, CIFAR-100 presents a more challenging task with 100 classes and fewer samples per class, requiring fine-grained feature discrimination. Furthermore, Tiny-ImageNet, a subset of ImageNet containing 200 classes with higher-resolution images ($64\times64$), introduces significantly higher dimensional complexity and semantic diversity.

Table~\ref{tab:dataset} summarizes the performance comparison. As observed, while baseline methods like FedAvg and SCAFFOLD suffer substantial performance degradation on the more complex CIFAR-100 and Tiny-ImageNet datasets, FedCoE maintains robust performance. Specifically, on Tiny-ImageNet, FedCoE outperforms the classic baselines and other MoE-based methods by a significant margin. This superiority is attributed to the specialized global expert pool, which effectively captures the diverse and complex semantic patterns inherent in large-scale datasets, validating the effectiveness of our consistency-driven aggregation strategy in handling complex tasks.

\begin{table}[t]
\centering
\setlength{\tabcolsep}{1.5mm} 
\renewcommand{\arraystretch}{1.1} 

\caption{Performance Comparison on Different Datasets (\%). The \textbf{best} results are highlighted in bold.}
\label{tab:dataset}

% \resizebox{\linewidth}{!}{
\small
    \begin{tabular}{lcccccc}
        \toprule
        \multirow{2}{*}{\textbf{Methods}} 
        & \multicolumn{2}{c}{\textbf{CIFAR-10}} 
        & \multicolumn{2}{c}{\textbf{CIFAR-100}} 
        & \multicolumn{2}{c}{\textbf{Tiny-ImageNet}} \\
        \cmidrule(lr){2-3} \cmidrule(lr){4-5} \cmidrule(lr){6-7}
        
        & $G_{acc}$ & $P_{acc}$ 
        & $G_{acc}$ & $P_{acc}$ 
        & $G_{acc}$ & $P_{acc}$ \\
        \midrule
        
        FedAvg\cite{mcmahan2017communication}   & 75.69 & 81.81 & 45.35 & 21.74 & 37.20 & 22.87 \\
        FedProx\cite{li2020federated}  & 80.05 & 86.52 & 46.36 & 49.45 & 36.94 & 36.56 \\
        SCAFFOLD\cite{karimireddy2020scaffold} & 44.41 & 35.84 & 41.47 & 28.05 & 30.00 & 23.35 \\
        FedMoEKD\cite{liang2025mixture} & -     & 56.86 & -     & 21.47 & -     &  8.75    \\
        pFedMoE\cite{yi2024pfedmoe}  & -     & 50.78 & -     & 14.70 & -     &  5.70    \\
        PM-MoE \cite{feng2025pm}  & 68.71 & 82.34 & 35.47 & 45.61 & 33.11     & 28.56     \\
        \midrule
        \textbf{FedCoE (Ours)} & \textbf{84.05} & \textbf{87.61} & \textbf{59.10} & \textbf{65.65} & \textbf{45.80} & \textbf{52.60} \\
        \bottomrule
    \end{tabular}
% }

\begin{tablenotes}
    \vspace{0.5em}
    \footnotesize
    \item \textit{Note: The symbol “-” indicates that $G_{acc}$ is not evaluated for methods without a unified global model. $G_{acc}$ and $P_{acc}$ denote Global and Personalized accuracy, respectively.}
\end{tablenotes}

\end{table}

\subsubsection{Cold-Start Generalization Capabilities}
To evaluate the practical applicability of FedCoE in dynamic environments, we mimic a realistic "cold-start" scenario involving the arrival of novel clients. Specifically, after convergence, three new clients with \textbf{unseen} data distributions are introduced. We evaluate performance under both moderate ($\alpha=0.5$) and extreme ($\alpha=0.1$) heterogeneity. 
For baselines (FedAvg, FedProx, Scaffold), we deploy the final aggregated global model for direct inference. For FedCoE, we leverage the \textit{Adaptive Cold-Start Mechanism}: new clients utilize the shared gating network to generate a semantic profile and retrieve a zero-shot personalized expert assembly from the server. Crucially, no local fine-tuning is performed, ensuring a strict evaluation of zero-shot generalization.

Table \ref{tab3:coldstart} reports the classification accuracy, yielding two critical insights:
\begin{itemize}
\item \textbf{Superior Zero-Shot Performance:} FedCoE achieves the highest average accuracy across all settings. Notably, under extreme heterogeneity ($\alpha=0.1$), it attains \textbf{91.27\%} accuracy, surpassing the strongest baseline (FedProx) by over \textbf{22\%}. This confirms that our correlation-aware expert assembly constructs highly tailored models for unseen distributions, overcoming the limitations of generic global models.
\item \textbf{Robustness to Heterogeneity:} While baseline performance deteriorates as heterogeneity increases (e.g., FedAvg drops from 64.73\% to 56.13\%), FedCoE displays an inverse trend, with gains becoming more pronounced at $\alpha=0.1$. This suggests that precise, correlation-based expert selection is particularly critical and effective in highly distinct heterogeneous environments.

\begin{table*}[t]
\centering
\setlength{\tabcolsep}{2.8mm} %2.2mm
\renewcommand{\arraystretch}{1}
\sethlcolor{color_pink}
\caption{Test accuracy (\%) on new clients (Cold-Start Scenario) under different non-IID settings ($\alpha=0.5$ and $\alpha=0.1$). ``Mean $\pm$ Std" denotes the average performance across the three new clients. The best results are highlighted in \textbf{bold}.}

\label{tab3:coldstart}
% \adjustbox{width=1\linewidth}{
{
\small
\begin{tabular}{lcccccccc}
    \toprule
    % --- 这是新的三行表头 ---
    
    % --- Row 1 ---
    \multirow{2}{*}{\textbf{Methods}} % Col 1: 跨2行    
    & \multicolumn{4}{c}{\textbf{$\alpha$ = 0.5}} % Col 1-4
    & \multicolumn{4}{c}{\textbf{$\alpha$ = 0.1}} % Col 5-8
    \\
    % Rules after Row 1
     \cmidrule{2-9}

    % --- Row 2 ---
    & \textbf{Client 1} 
    & \textbf{Client 2} 
    & \textbf{Client 3} & \textbf{$\mu + \sigma$} 
    & \textbf{Client 1} 
    & \textbf{Client 2} 
    & \textbf{Client 3} & \textbf{$\mu + \sigma$} 
     \\ % Col 13 (spacer)

    % --- 表头结束 ---
%%%%%%%%%%%%%% 
\midrule[0.5pt]      
FedAvg\cite{mcmahan2017communication} & 55.00 & 73.40 & 65.80 & 64.73 $\pm$ 9.25 & 69.00 & 46.60 & 52.80 & 56.13 $\pm$ 11.57 \\
FedProx\cite{li2020federated} & 75.00 & 79.00 & 74.20 & 76.07 $\pm$ 2.57 & 73.60 & 73.20 & 60.60 & 69.13 $\pm$ 7.39 \\
Scaffold\cite{karimireddy2020scaffold} & 18.00 & 33.40 & 31.80 & 27.73 $\pm$ 8.47 & 27.80 & 7.60 & 9.20 & 14.87 $\pm$ 11.23 \\
\textbf{FedCoE (Ours)} & \textbf{76.80} & \textbf{80.00} & \textbf{75.00} & \textbf{77.27 $\pm$ 2.53} & \textbf{89.80} & \textbf{90.60} & \textbf{93.40} & \textbf{91.27 $\pm$ 1.89} \\

 \bottomrule
       \end{tabular}
% }
}

\end{table*}

\item \textbf{Stability:} As evidenced by the standard deviation ($\sigma$), FedCoE exhibits the lowest performance variance. Whether the new client's distribution is balanced or skewed, our mechanism ensures a reliable initialization. In contrast, methods like SCAFFOLD show extreme instability (e.g., dropping to 7.6\% accuracy), likely due to the misalignment of historical control variates with the new client's data.
\end{itemize}

\begin{figure}[]
    \centering
    \includegraphics[width=1\linewidth]{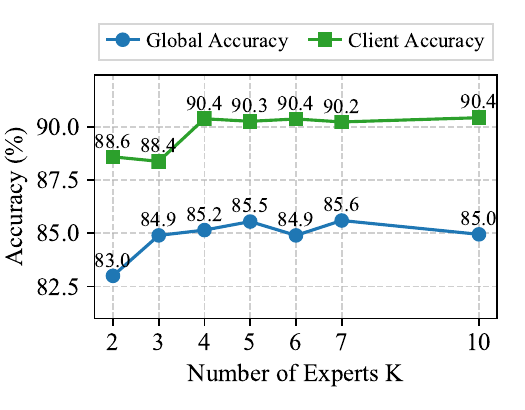}
    \caption{The impact of the number of experts (K) on model performance.}
    \label{fig:e_num}
\end{figure}

\subsubsection{Ablation Studies}
\begin{figure}[]
    \centering
    \includegraphics[width=1\linewidth]{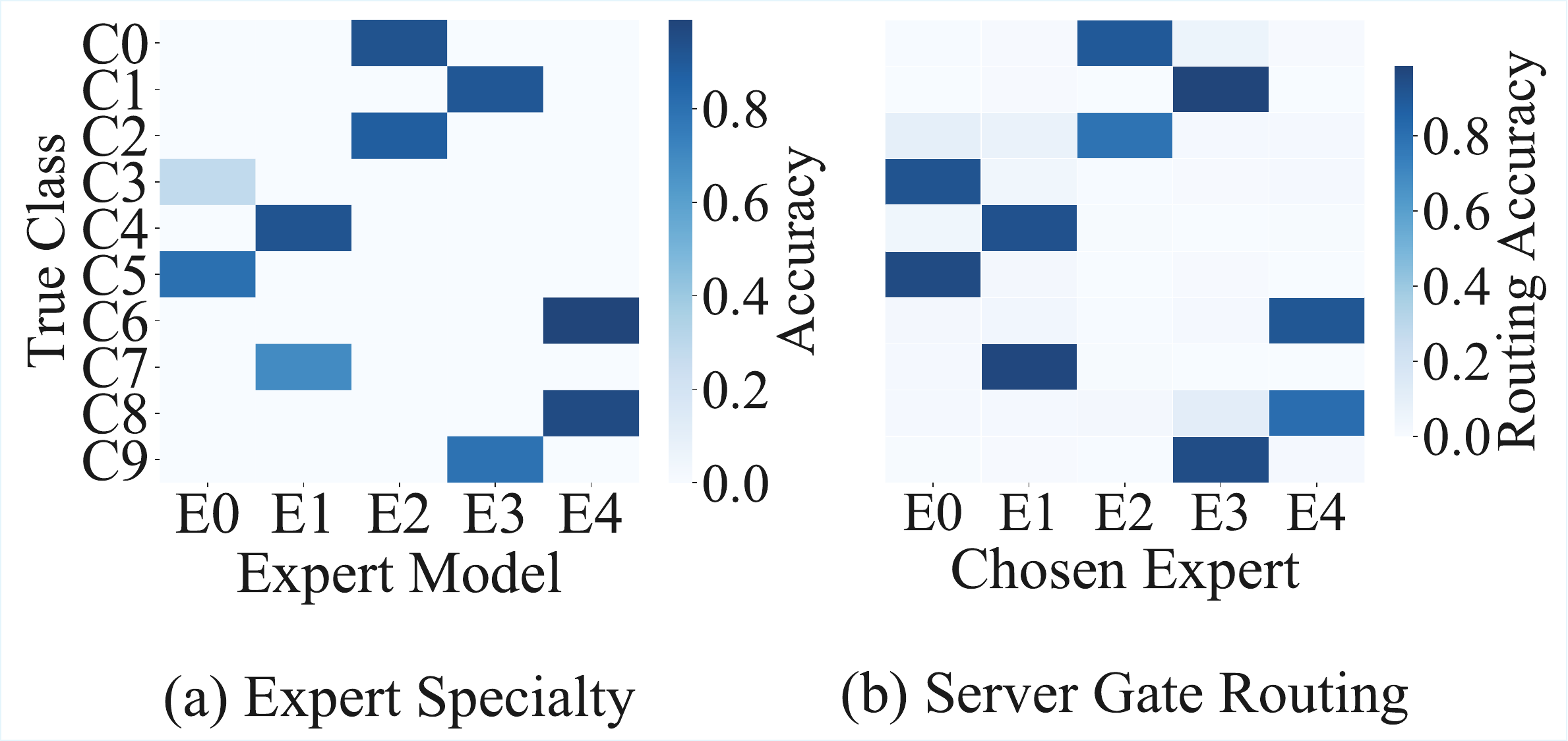}
    \caption{Validation of the gate network's selection effectiveness}
    \label{fig:gate}
\end{figure}

\begin{itemize}
\item \textbf{Impact of Core Modules:}
We perform a module-level ablation study (Table~\ref{tab:module_ablation}) to quantify component contributions. A single-expert baseline yields 80.0\% accuracy. Simply increasing expert count without a routing mechanism degrades performance to 78.0\%, indicating that uncoordinated experts introduce noise. Integrating the gating network leverages the multi-expert capacity, boosting accuracy to 89.0\%. This confirms that adaptive routing is indispensable for coordinating expert specialization and unlocking the potential of the ensemble.

\item \textbf{Sensitivity to Expert Count ($K$):}
We analyze the impact of expert pool size, varying $K$ from 2 to 10 (Fig~\ref{fig:e_num}). Accuracy improves significantly as $K$ increases to 4, reflecting enhanced capacity to model diverse patterns. Beyond $K=4$, performance saturates at $\approx 90.4\%$ and remains stable up to $K=10$, demonstrating the robustness of our framework. However, larger $K$ expands the optimization search space, slowing convergence. Consequently, we identify $K \in [4, 5]$ as the optimal range balancing model capacity and computational efficiency.

% \item[$\bullet$] \textbf{Gating effectiveness}
% To further examine the gating mechanism in FedCoE, we visualize the expert specialization and the gating assignments across classes. As shown in Figure~\ref{fig:gate}, the first heatmap depicts the distribution of expert specialization, where each expert demonstrates higher activation on a subset of classes, indicating distinct domain preferences. The second heatmap shows the probabilities that the gating network assigns to each expert for each class. Remarkably, the gating pattern aligns closely with the experts’ learned specialization — the gating network consistently assigns samples to the experts that perform best for the corresponding classes. This strong correlation demonstrates that the FedCoE gating module effectively captures inter class relationships and dynamically routes clients to the most relevant experts, leading to more efficient and stable personalization.

\item \textbf{Interpretability of Expert Specialization:}
To verify that our correlation-aware framework is grounded in meaningful semantics, we visualize expert behavior for $K=5$. Fig.~\ref{fig:gate}(a) depicts the validation accuracy of individual experts, revealing distinct specialization patterns. Fig.~\ref{fig:gate}(b) shows the aggregate routing probabilities assigned by the gating network.
A strong semantic alignment is evident: the gating network consistently routes samples to experts with the highest empirical performance for their respective classes. This correlation confirms that the gating module captures genuine class-level semantics. It validates the reliability of our Client-Expert Correlation Matrix, ensuring that aggregation is driven by functional competence rather than stochastic noise.
\end{itemize}

% \begin{table}[t]
% \centering
% \caption{Module-level ablation analysis of FedCoE.}
% \label{tab:module_ablation}
% {
% \small
% \begin{tabular}{lc}
% \toprule
% \textbf{Methods} & \textbf{Accuracy (\%)} \\
% \midrule
% Single Expert(FedAvg) & 81.81 \\
% + Multiple Experts & 79.52 \\
% + Multiple Experts + Gate  & \textbf{88.57} \\
% \bottomrule
% \end{tabular}
% }
% \end{table}

\begin{table}[t]
\centering
\caption{Module-level ablation analysis of FedCoE.}
\label{tab:module_ablation}
{
\small
\begin{tabular}{lc}
\toprule
\textbf{Methods} & \textbf{Accuracy (\%)} \\
\midrule
w/o Multiple Experts \& Gate & 81.81 \\
w/o Gate & 79.52 \\
\textbf{FedMoE(ours)}  & \textbf{88.57} \\
\bottomrule
\end{tabular}
}
\end{table}

\section{Conclusion}\label{sec:conclusion}
In this paper, we presented FedCoE, a federated coordinated dual-level mixture-of-experts framework that bridges the gap between global generalization and local personalization in non-IID federated learning. The core innovation lies in the synergy of a shared semantic gating network and a consistency-driven expert aggregation strategy, which jointly address expert drift and gating inconsistency that have limited prior MoE-based FL approaches. Extensive experiments demonstrate that FedCoE achieves 78.00\% global accuracy and 89.32\% personalized accuracy, while delivering 77.27\% accuracy in cold-start scenarios without any local fine-tuning.
% In this paper, we presented \textbf{FedCoE}, a correlation-aware dual-expert framework that successfully bridges the gap between global generalization and local personalization in non-IID Federated Learning. Our core contribution lies in the synergy of a \textit{shared server-side gating network} and a \textit{correlation-weighted aggregation mechanism}. This innovative design addresses the fundamental challenges of expert drift and gating inconsistency, enabling the precise filtering and integration of specialized knowledge from diverse clients. Consequently, FedCoE not only achieves superior performance in highly heterogeneous environments but also offers a pioneering \textbf{zero-shot cold-start solution} that enables instant adaptation to new clients without local fine-tuning.

Looking forward, we identify several promising directions to extend this work. \textit{a) Multi-task Federated Learning}: We plan to adapt FedCoE for multi-task federated scenarios, exploring how dynamic routing can facilitate positive knowledge transfer across distinct but related tasks. \textit{b) Resource-efficient Expert Activation}: We aim to develop lightweight expert activation strategies that enable the deployment of large-scale MoE models on edge devices with strict memory and computation constraints. \textit{(c) Privacy-preserving Mechanisms}: Integrating differential privacy into gating signal transmission remains a critical direction to further protect client semantic profiles during the correlation modeling process.
% Looking forward, we identify several promising avenues to extend this work. First, we plan to adapt FedCoE for \textbf{multi-task federated learning}, exploring how dynamic routing can facilitate positive knowledge transfer across distinct but related tasks. Second, we aim to investigate \textbf{resource-efficient expert activation} strategies to deploy large-scale MoE models on edge devices with strict memory and computation constraints. Finally, integrating \textbf{privacy-preserving mechanisms}, such as Differential Privacy, into the gating signal transmission to further protect client semantic profiles remains a critical direction for future research.

% \section*{Acknowledgment}

\bibliographystyle{IEEEtran}
\bibliography{refs}

@article{arivazhagan2019federated,
  author = {Manoj Ghuhan Arivazhagan and Vinay Aggarwal and Aaditya Kumar Singh and Sunav Choudhary},
  journal = {arXiv preprint arXiv:1912.00818},
  title = {Federated Learning with Personalization Layers},
  year = {2019}
}

@inproceedings{deng2009imagenet,
  author = {Jia Deng and Wei Dong and Richard Socher and Li-Jia Li and Kai Li and Li Fei-Fei},
  booktitle = {Proceedings of IEEE Conference on Computer Vision and Pattern Recognition},
  pages = {248--255},
  title = {Imagenet: a Large-scale Hierarchical Image Database},
  year = {2009}
}

@article{dunfedjets,
  author={Yehya Farhat and Hamza ElMokhtar Shili and Fangshuo Liao and Chen Dun and Mirian Hipolito Garcia and Guoqing Zheng and Ahmed Hassan Awadallah and Robert Sim and Dimitrios Dimitriadis and Anastasios Kyrillidis},
  journal = {arXiv preprint arXiv:2306.08586},
  title={Learning to Specialize: Joint Gating-Expert Training for Adaptive MoEs in Decentralized Settings}, 
  year = {2025}
}

@article{fallah2020personalized,
  author = {Alireza Fallah and Aryan Mokhtari and Asuman Ozdaglar},
  journal = {arXiv preprint arXiv:2002.07948},
  title = {Personalized Federated Learning: a Meta-learning Approach},
  year = {2020}
}

@article{fedus2022switch,
  author = {William Fedus and Barret Zoph and Noam Shazeer},
  journal = {Journal of Machine Learning Research},
  number = {120},
  pages = {1--39},
  title = {Switch Transformers: Scaling to Trillion Parameter Models with Simple and Efficient Sparsity},
  volume = {23},
  year = {2022}
}

@inproceedings{feng2025pm,
  author = {Yu Feng and Yangli-ao Geng and Yifan Zhu and Zongfu Han and Xie Yu and Kaiwen Xue and Haoran Luo and Mengyang Sun and Guangwei Zhang and Meina Song},
  booktitle = {Proceedings of ACM on Web Conference},
  pages = {134--146},
  title = {{PM-MOE}: Mixture of Experts on Private Model Parameters for Personalized Federated Learning},
  year = {2025}
}

@inproceedings{jiang2025heterogeneous,
  author = {Jingang Jiang and Yanzhao Chen and Xiangyang Liu and Haiqi Jiang and Chenyou Fan},
  booktitle = {Proceedings of International Joint Conference on Artificial Intelligence},
  pages = {5480--5488},
  title = {Heterogeneous Federated Learning with Scalable Server Mixture-of-experts},
  year = {2025}
}

@article{kairouz2021advances,
  author = {Peter Kairouz and H Brendan McMahan and Brendan Avent and Aurélien Bellet and Mehdi Bennis and Arjun Nitin Bhagoji and Kallista Bonawitz and Zachary Charles and Graham Cormode and Rachel Cummings and others},
  journal = {Foundations and trends{\textregistered} in machine learning},
  number = {1--2},
  pages = {1--210},
  publisher = {Now Publishers, Inc.},
  title = {Advances and Open Problems in Federated Learning},
  volume = {14},
  year = {2021}
}

@inproceedings{karimireddy2020scaffold,
  author = {Sai Praneeth Karimireddy and Satyen Kale and Mehryar Mohri and Sashank Reddi and Sebastian Stich and Ananda Theertha Suresh},
  booktitle = {Proceedings of International Conference on Machine Learning},
  organization = {PMLR},
  pages = {5132--5143},
  title = {Scaffold: Stochastic Controlled Averaging for Federated Learning},
  year = {2020}
}

@article{krizhevsky2009learning,
  author = {Alex Krizhevsky and Geoffrey Hinton and others},
  publisher = {Toronto, ON, Canada},
  title = {Learning Multiple Layers of Features from Tiny Images},
  year = {2009}
}

@article{lecun2015deep,
  author = {Yann LeCun and Yoshua Bengio and Geoffrey Hinton},
  journal = {Nature},
  number = {7553},
  pages = {436--444},
  publisher = {Nature Publishing Group UK London},
  title = {Deep Learning},
  volume = {521},
  year = {2015}
}

@article{lepikhin2020gshard,
  author = {Dmitry Lepikhin and HyoukJoong Lee and Yuanzhong Xu and Dehao Chen and Orhan Firat and Yanping Huang and Maxim Krikun and Noam Shazeer and Zhifeng Chen},
  journal = {arXiv preprint arXiv:2006.16668},
  title = {Gshard: Scaling Giant Models with Conditional Computation and Automatic Sharding},
  year = {2020}
}

@inproceedings{li2020federated,
  author = {Tian Li and Anit Kumar Sahu and Manzil Zaheer and Maziar Sanjabi and Ameet Talwalkar and Virginia Smith},
  booktitle = {Proceedings of Annual Conference on Machine Learning and System},
  pages = {429--450},
  title = {Federated Optimization in Heterogeneous Networks},
  volume = {2},
  year = {2020}
}

@inproceedings{li2022federated,
  author = {Qinbin Li and Yiqun Diao and Quan Chen and Bingsheng He},
  booktitle = {Proceedings of IEEE International Conference on Data Engineering},
  organization = {IEEE},
  pages = {965--978},
  title = {Federated Learning on Non-iid Data Silos: an Experimental Study},
  year = {2022}
}

@article{liang2020thinklocallyactglobally,
  author = {Paul Pu Liang and Terrance Liu and Liu Ziyin and Nicholas B. Allen and Randy P. Auerbach and David Brent and Ruslan Salakhutdinov and Louis-Philippe Morency},
  journal = {arXiv preprint arXiv:2001.01523},
  title = {Think Locally, Act Globally: Federated Learning with Local and Global Representations},
  year = {2020}
}

@article{liang2025mixture,
  author = {Tonghui Liang and Meng Hu and Enchang Sun},
  journal = {IEEE Networking Letters },
  publisher = {IEEE},
  title = {Mixture of Specialized Experts for {Model-Heterogeneous} Personalized Federated Learning},
  year={2025},
  volume={7},
  number={3},
  pages={224-228}
}

@inproceedings{m2024personalized,
  author = {Pouya M Ghari and Yanning Shen},
  booktitle = {Proceedings of Advances in Neural Information Processing Systems},
  pages = {92155--92183},
  title = {Personalized Federated Learning with Mixture of Models for Adaptive Prediction and Model Fine-tuning},
  year = {2024}
}

@inproceedings{mcmahan2017communication,
  author = {Brendan McMahan and Eider Moore and Daniel Ramage and Seth Hampson and Blaise Aguera y Arcas},
  booktitle = {Proceedings of International Conference on Artificial Intelligence and Statistics},
  organization = {PMLR},
  pages = {1273--1282},
  title = {Communication-efficient Learning of Deep Networks from Decentralized Data},
  year = {2017}
}

@article{mei2024fedmoe,
  author = {Hanzi Mei and Dongqi Cai and Ao Zhou and Shangguang Wang and Mengwei Xu},
  journal = {arXiv preprint arXiv:2408.11304},
  title = {Fedmoe: Personalized Federated Learning via Heterogeneous Mixture of Experts},
  year = {2024}
}

@article{pardau2018california,
  author = {Stuart L Pardau},
  journal={J. Tech. L. \& Pol'y},
  pages = {68},
  publisher = {HeinOnline},
  title = {The California Consumer Privacy Act: towards a European-style Privacy Regime in the United States},
  volume = {23},
  year = {2018}
}

@inproceedings{rajbhandari2022deepspeed,
  author = {Samyam Rajbhandari and Conglong Li and Zhewei Yao and Minjia Zhang and Reza Yazdani Aminabadi and Ammar Ahmad Awan and Jeff Rasley and Yuxiong He},
  booktitle = {Proceedings of International Conference on Machine Learning},
  organization = {PMLR},
  pages = {18332--18346},
  title = {Deepspeed-moe: Advancing Mixture-of-experts Inference and Training to Power Next-generation {ai} Scale},
  year = {2022}
}

@article{regulation2018general,
  author = {Protection Regulation},
  journal = {Intouch},
  pages = {1--5},
  title = {General Data Protection Regulation},
  volume = {25},
  year = {2018}
}

@inproceedings{riquelme2021scaling,
  author = {Carlos Riquelme and Joan Puigcerver and Basil Mustafa and Maxim Neumann and Rodolphe Jenatton and André Susano Pinto and Daniel Keysers and Neil Houlsby},
  booktitle = {Proceedings of Advances in Neural Information Processing Systems},
  pages = {8583--8595},
  title = {Scaling Vision with Sparse Mixture of Experts},
  volume = {34},
  year = {2021}
}

@inproceedings{scott2024pefll,
  author = {Jonathan Scott and Hossein Zakerinia and Christoph H Lampert},
  booktitle = {Proceedings of International Conference on Learning Representations},
  title = {{PeFLL}: Personalized Federated Learning by Learning to Learn},
  year = {2024}
}

@article{shazeer2017outrageously,
  author = {Noam Shazeer and Azalia Mirhoseini and Krzysztof Maziarz and Andy Davis and Quoc Le and Geoffrey Hinton and Jeff Dean},
  journal = {arXiv preprint arXiv:1701.06538},
  title = {Outrageously Large Neural Networks: the Sparsely-gated Mixture-of-experts Layer},
  year = {2017}
}

@inproceedings{wang2023towards,
  author = {Jiaqi Wang and Xingyi Yang and Suhan Cui and Liwei Che and Lingjuan Lyu and Dongkuan DK Xu and Fenglong Ma},
  booktitle = {Proceedings of Advances in Neural Information Processing Systems},
  pages = {29515--29531},
  title = {Towards Personalized Federated Learning via Heterogeneous Model Reassembly},
  volume = {36},
  year = {2023}
}

@article{wu2022communication,
  author = {Chuhan Wu and Fangzhao Wu and Lingjuan Lyu and Yongfeng Huang and Xing Xie},
  journal = {Nature Communications},
  number = {1},
  pages = {2032},
  publisher = {Nature Publishing Group UK London},
  title = {Communication-efficient Federated Learning via Knowledge Distillation},
  volume = {13},
  year = {2022}
}

@inproceedings{wu2024fedmoe,
  author = {Chuan Wu},
  booktitle = {Proceedings of International Conference on Mobility, Sensing and Networking},
  title = {{FedMoE-DA}: Federated Mixture of Experts via Domain Aware Fine-grained Aggregation},
  year = {2024}
}

@article{xu2024overcoming,
  author = {Yang Xu and Yunming Liao and Lun Wang and Hongli Xu and Zhida Jiang and Wuyang Zhang},
  journal = {IEEE Transactions on Mobile Computing},
  number = {12},
  pages = {11406--11421},
  publisher = {IEEE},
  title = {Overcoming Noisy Labels and Non-iid Data in Edge Federated Learning},
  volume = {23},
  year = {2024}
}

@inproceedings{yao2024perfedrlnas,
  author = {Dixi Yao and Baochun Li},
  booktitle = {Proceedings of AAAI Conference on Artificial Intelligence},
  number = {15},
  pages = {16398--16406},
  title = {{PerFedRLNAS}: One-for-all Personalized Federated Neural Architecture Search},
  volume = {38},
  year = {2024}
}

@article{yi2024pfedmoe,
  author = {Liping Yi and Han Yu and Chao Ren and Heng Zhang and Gang Wang and Xiaoguang Liu and Xiaoxiao Li},
  journal = {arXiv preprint arXiv:2402.01350},
  title = {Pfedmoe: Data-level Personalization with Mixture of Experts for Model-heterogeneous Personalized Federated Learning},
  year = {2024}
}

@inproceedings{zhang2023robust,
  author = {Yihua Zhang and Ruisi Cai and Tianlong Chen and Guanhua Zhang and Huan Zhang and Pin-Yu Chen and Shiyu Chang and Zhangyang Wang and Sijia Liu},
  booktitle = {Proceedings of IEEE/CVF International Conference on Computer Vision},
  pages = {90--101},
  title = {Robust Mixture-of-expert Training for Convolutional Neural Networks},
  year = {2023}
}

@inproceedings{zhangpersonalized,
  author = {Michael Zhang and Karan Sapra and Sanja Fidler and Serena Yeung and Jose M Alvarez},
  booktitle = {Proceedings of International Conference on Learning Representations},
  title = {Personalized Federated Learning with First Order Model Optimization},
  year = {2021}
}

@inproceedings{zoph2022designing,
  author = {Barret Zoph},
  booktitle = {Proceedings of IEEE International Parallel and Distributed Processing Symposium Workshops},
  organization = {IEEE},
  pages = {1044--1044},
  title = {Designing Effective Sparse Expert Models},
  year = {2022}
}

@inproceedings{zou2024fed,
  author = {Yifei Zou and Senmao Qi and Yuan Yuan and Dawei Wang and Shikun Shen and Libing Wu and Shaoyong Guo and Dongxiao Yu},
  booktitle = {Proceedings of International Conference on Parallel and Distributed Computing: Applications and Technologies},
  organization = {Springer},
  pages = {128--139},
  title = {{Fed-MoE}: Efficient Federated Learning for {Mixture-of-Experts} Models via Empirical Pruning},
  year = {2024}
}

@inproceedings{paszke2019pytorch,
  title={PyTorch: An Imperative Style, High-Performance Deep Learning Library},
  author={Paszke, Adam and Gross, Sam and Massa, Francisco and Lerer, Adam and Bradbury, James and Chanan, Gregory and Killeen, Trevor and Lin, Zeming and Gimelshein, Natalia and Antiga, Luca and others},
  booktitle={Proceedings of Advances in Neural Information Processing Systems},
  volume={32},
  year={2019}
}

\begin{IEEEbiography}[{\includegraphics[width=1\textwidth]{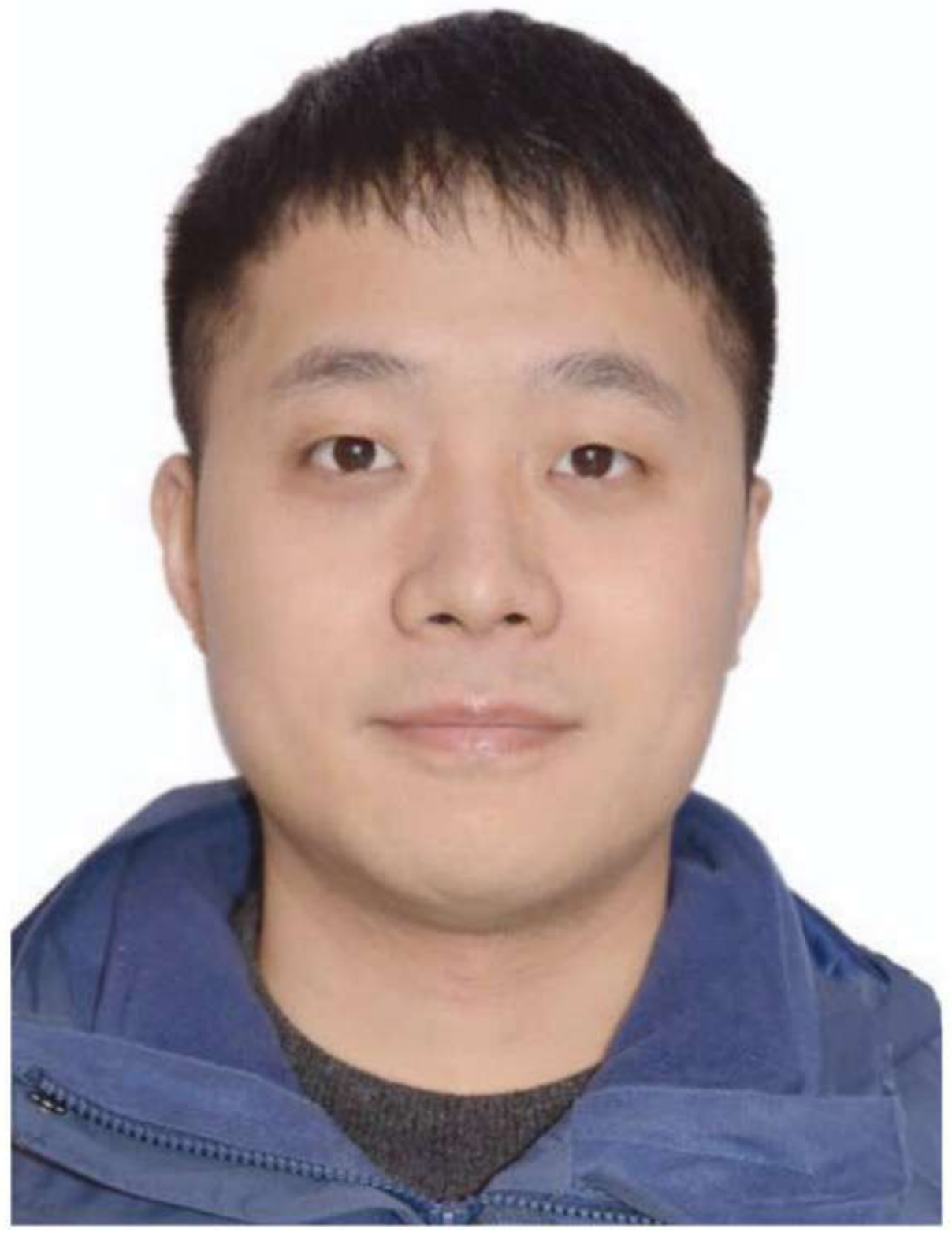}}]{Penglin Dai}(S'15-M'17) received the B.S. degree in mathematics and applied mathematics and the Ph.D. degree in computer science from Chongqing University, Chongqing, China, in 2012 and 2017, respectively. He is currently an Associate Professor with the School of Computing and Artificial Intelligence, Southwest Jiaotong University, Chengdu, China. His research interests include internet of vehicles, wireless networks and mobile computing.
\end{IEEEbiography}

\begin{IEEEbiography}[{\includegraphics[width=1\textwidth]{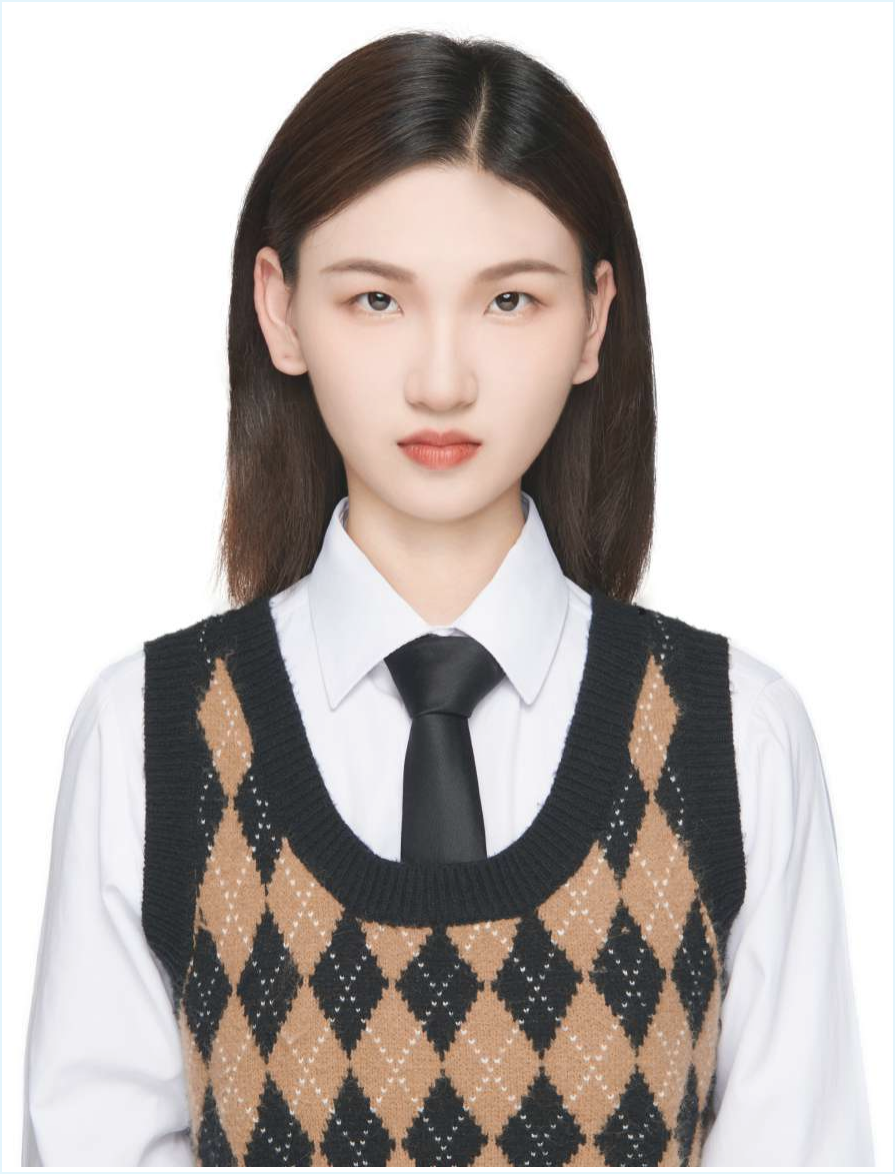}}]{Fulian Li} received the B.E. degree in Computer Science and Technology from Chengdu University of Technology, Chengdu, China, in 2024. She is currently pursuing the master's degree in Computer Science and Technology of Southwest Jiaotong University. Her research interests include federated learning in edge computing and mixture of experts.
\end{IEEEbiography}

\begin{IEEEbiography}[{\includegraphics[width=1in,height=1.25in,clip,keepaspectratio]{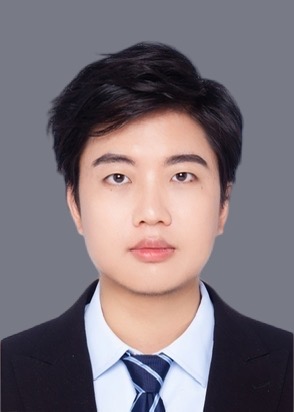}}]{Xincao Xu}
(Member, IEEE) received the B.E. degree from the North University of China, Taiyuan, China, in 2017, and the Ph.D. degree from the Chongqing University, Chongqing, China, in 2023. He is currently an Associate Researcher in computer science with the Shenzhen Institute for Advanced Study, University of Electronic Science and Technology of China (UESTC), Shenzhen, China. From 2023 to 2025, he was a Postdoctoral Research Fellow with the Shenzhen Institute for Advanced Study, UESTC. His research interests include edge intelligence, agentic artificial intelligence, and agentic reinforcement learning.
\end{IEEEbiography}

\begin{IEEEbiography}[{\includegraphics[width=1\textwidth]{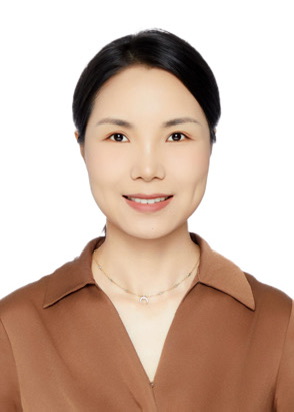}}]{Junhua Wang} (Member, IEEE) received the BS degree and PhD degree in computer science from Chongqing University, in 2014, and 2019, respectively. From 2017 to 2018, she was a visiting scholar in University of Houston, USA. Now she is an associate professor of the school of computer science and engineering, Northeastern University. Her research interests include mobile computing, internet of vehicles, and wireless networks.
\end{IEEEbiography}

\begin{IEEEbiography}[{\includegraphics[width=1\textwidth]{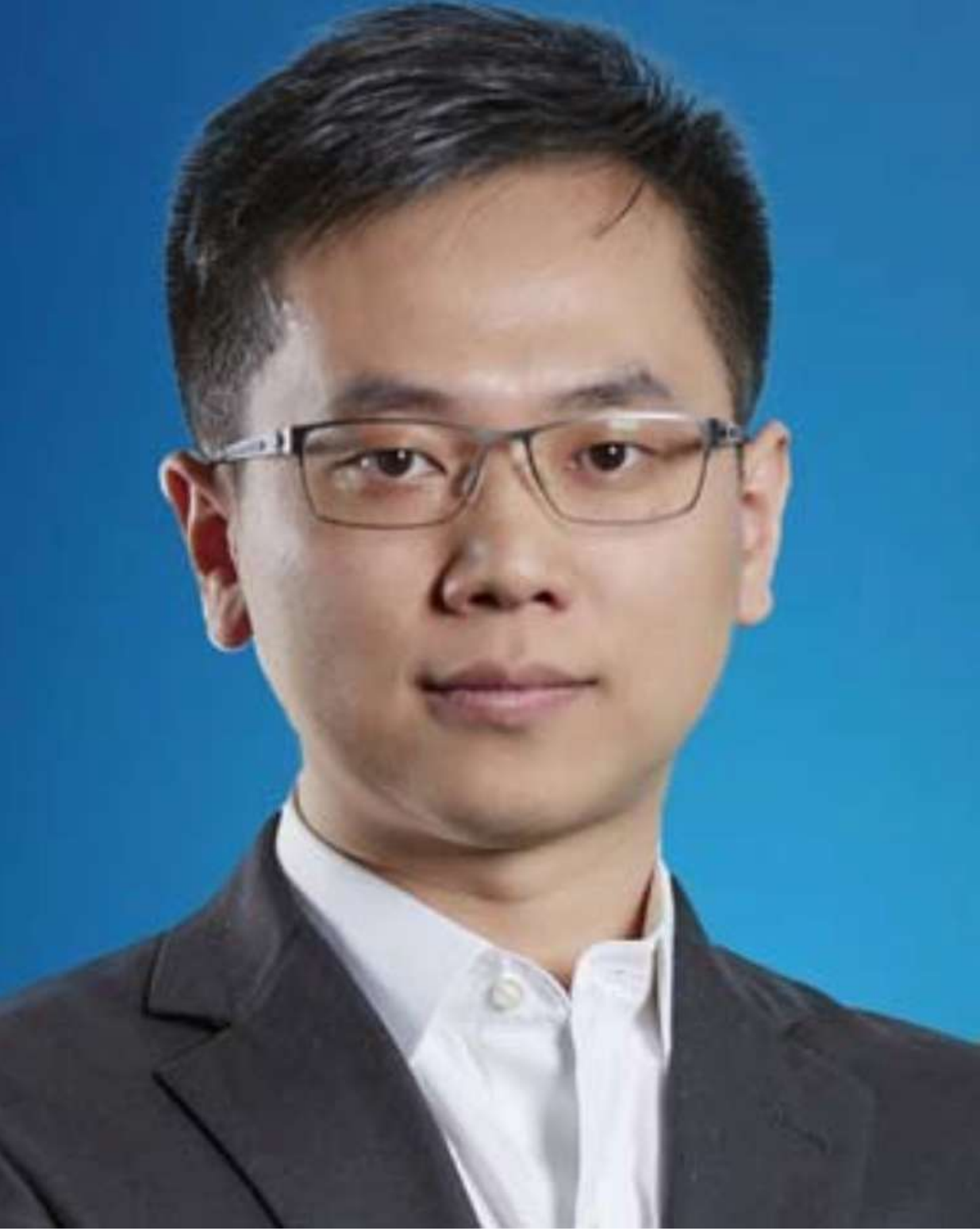}}]{Lixin Duan} ((Member, IEEE) received the B.E. degree from the University of Science and Technology of China, Hefei, China, in 2008, and the Ph.D. degree from Nanyang Technological University, Singapore, in 2012. He is currently a Full Professor with the University of Electronic Science and Technology of China. His current research interests include transfer learning, multiple instance learning, and their applications in computer vision and data mining.
\end{IEEEbiography}

% \begin{IEEEbiography}[{\includegraphics[width=1\textwidth]{Huanlai_Xing.pdf}}]{Huanlai Xing} (M'16) received his B. Eng. degree in communications engineering from Southwest Jiaotong University, Chengdu, China, in 2006; his M.Sc. degree in electromagnetic fields and wavelength technology from Beijing University of Posts and Telecommunications, Beijing, China, in 2009; and his Ph.D. degree in computer science from University of Nottingham, Nottingham, U.K., in 2013. He is an Associate Professor with the School of Computing and Artificial Intelligence, Southwest Jiaotong University. His research interests include  network function virtualization, software defined networks, machine learning, data mining, evolutionary computation, multiobjective optimization, etc. He has authored and co-authored over 60 peer-reviewed journal and conference papers. Dr. Xing is a Member of IEEE.
% \end{IEEEbiography}

\begin{IEEEbiography}[{\includegraphics[width=1\textwidth]{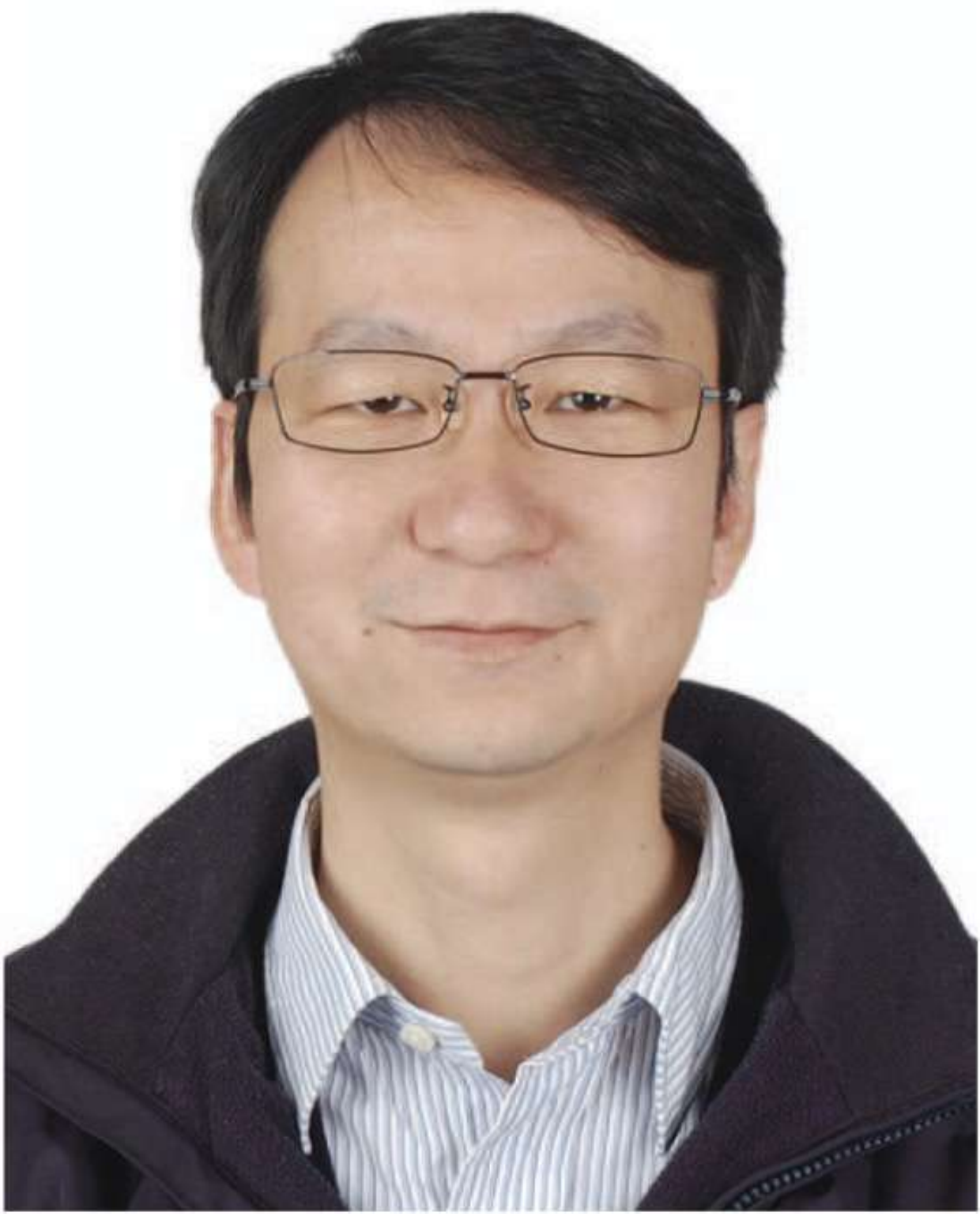}}]
	{Xiao Wu} (S'05-M'08) received the B.Eng. and M.S. degrees in computer science from Yunnan University, Yunnan, China, in 1999 and 2002, respectively, and the Ph.D. degree in Computer Science from City University of Hong Kong, Hong Kong in 2008.
	Currently, he is a Professor and the Assistant Dean of School of Computing and Artificial Intelligence, Southwest Jiaotong University, Chengdu, China. He was with the Institute of Software, Chinese Academy of Sciences, Beijing, China, from 2001 to 2002. He was a Research Assistant and a Senior Research Associate at the City University of Hong Kong, Hong Kong, from 2003 to 2004, and 2007 to 2009, respectively. He was with the School of Computer Science, Carnegie Mellon University, Pittsburgh, PA, USA, and at School of Information and Computer Science, University of California, Irvine, CA, USA as a Visiting Scholar during 2006 to 2007 and 2015 to 2016, respectively. He has authored or co-authored more than 100 research papers in well-respected journals, such as TIP, TMM, TMI and prestigious proceedings like CVPR, ICCV and ACM MM. He received the Second Prize of Natural Science Award of the Ministry of Education, China in 2016 and the Second Prize of Science and Technology Progress Award of Henan Province, China in 2017. His research interests include artificial intelligence, computer vision, multimedia information retrieval, and image/video computing.
\end{IEEEbiography}

\end{document}